\documentclass{article}

\usepackage[main, final]{neurips_2026}

\usepackage[utf8]{inputenc} 
\usepackage[T1]{fontenc}    
\usepackage{hyperref}       
\usepackage{url}            
\usepackage{booktabs}       
\usepackage{amsfonts}       
\usepackage{nicefrac}       
\usepackage{microtype}      
\usepackage{xcolor}         

\usepackage{graphicx}    
\usepackage{multirow}    
\usepackage{colortbl}    
\usepackage[dvipsnames]{xcolor}
\usepackage{amsmath}
\usepackage{wrapfig}
\usepackage{caption}
\usepackage{enumitem}
\usepackage{algorithm}
\usepackage{algpseudocode}

\definecolor{pubcolor}{RGB}{88, 100, 160}
\definecolor{pubcolor_conference}{RGB}{60, 70, 115}
\newcommand{\pub}[1]{$({\small\textcolor{pubcolor_conference}{\text{#1}}})$}

\definecolor{rowblue}{RGB}{239, 245, 251}

\title{Towards Generalizable 3D Anomaly Detection \\ via Relational Inconsistency Modeling}

\author{%
  KunHo Heo\footnotemark[1] \\
  Kyung Hee University\\
  \texttt{hkh7710@khu.ac.kr} \\
  \And
  SuYeon Kim\footnotemark[1] \\
  Kyung Hee University\\
  \texttt{spoiuy3@khu.ac.kr} \\
  \AND
  Hayoung Lee \\
  Kyung Hee University\\
  \texttt{lhayoung9@khu.ac.kr} \\
  \And
  Chanse Oh \\
  Kyung Hee University\\
  \texttt{chanse0727@khu.ac.kr} \\
  \And
  MyeongAh Cho\footnotemark[2] \\
  Kyung Hee University\\
  \texttt{maycho@khu.ac.kr} \\
}
\renewcommand{\thefootnote}{\fnsymbol{footnote}}

\begin{document}

\maketitle

\begingroup
\renewcommand\thefootnote{}
\footnotetext{* Equal contribution}
\footnotetext{\dag\ Corresponding author}
\endgroup

\begin{abstract}
3D anomaly detection (3DAD) aims to identify defective regions in point cloud data, serving as a critical component in industrial inspection systems. Existing methods are normality-centered — learning the distribution of normal samples and treating deviations as anomalies — without explicitly modeling what constitutes a defect. This leads to ambiguous decision boundaries with increased false positives and negatives, particularly in unified and cross-domain settings where diverse normal distributions further blur the boundaries. We propose a \textbf{relational inconsistency} modeling framework that characterizes defects as violations of geometric consistency among neighboring structures. Our approach learns category-agnostic defect cues through pseudo-anomalies designed as controlled relational violations, instantiated by two key modules: \textbf{Edge-aware Graph Refinement (EGR)} for encoding geometric relationships among local regions, and \textbf{Cluster-Deviation Modeling (CDM)} for identifying regions that are relationally incompatible within their structural peer group. Extensive experiments on Anomaly-ShapeNet and Real3D-AD demonstrate consistent improvements over prior state-of-the-art methods in both in-domain and cross-domain settings, validating the effectiveness of learning an explicit, relation-based defect criterion for 3D anomaly detection. Project page: \href{https://visualsciencelab-khu.github.io/GRIM_project/}{\textcolor{magenta}{https://github.com/VisualScienceLab-KHU/GRIM}}.
\end{abstract}
\section{Introduction}
3D anomaly detection (3DAD) aims to identify defective regions in point cloud data and serves as a critical component in industrial inspection and automation systems. 
Unlike anomaly detection in 2D images~\cite{yi2020patch, defard2021padim, zavrtanik2021reconstruction, ristea2022self}, which primarily relies on appearance cues such as color and texture, defects in 3D environments frequently manifest as violations of geometric consistency among spatially neighboring structures rather than as absolute abnormalities of isolated local regions.
Due to the inherent scarcity of anomalous samples in real-world industrial settings, 3DAD is typically formulated as an unsupervised problem, where a model learns the distribution of normality from defect-free data and identifies deviations as anomalies at test time~\cite{liu2023real3d, zhu2024towards, li2024towards, zhou2024r3d, ijcai2025p94, liang2025taming}.

\begin{figure}[t]
    \centering
    \includegraphics[width=1\linewidth]{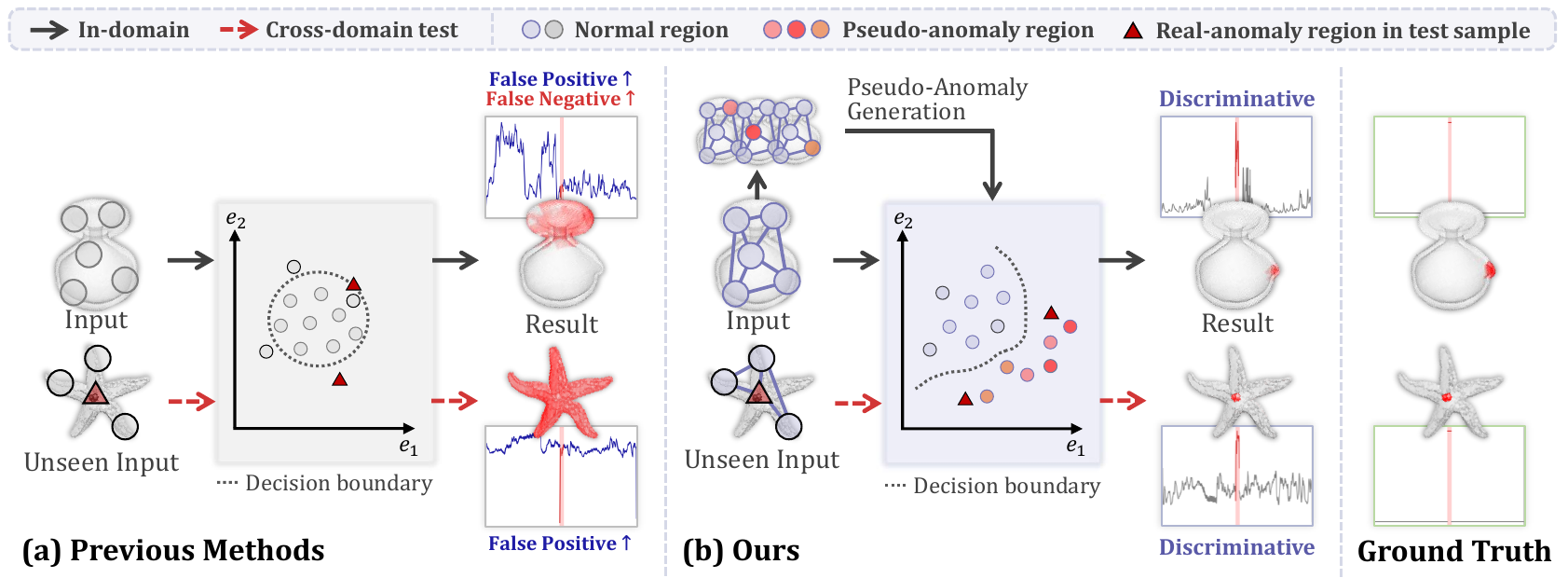}
    \caption{Comparison of previous normality-centered 3DAD methods and our relation-based defect learning framework. 
    (a) Previous methods (e.g., MC3D-AD) rely on normal feature distributions, often resulting in false positives and false negatives. 
    (b) Our method uses pseudo-anomalies as controlled violations of geometric relations among normal structures. 
    By distinguishing relation-violating pseudo-anomalies from normal regions, the model produces more discriminative point-level anomaly scores, reducing both false positives and false negatives.}
    \label{fig:figure1}
\end{figure}

Prior work on 3DAD has largely evolved along two main directions: reconstruction-based methods and feature-similarity-based methods. 
Reconstruction-based methods train a model to reconstruct normal samples and use reconstruction error as an anomaly score, under the assumption that anomalous regions will fail to be reconstructed accurately. Feature-similarity-based methods, often implemented via memory banks, determine abnormality by measuring how dissimilar a test feature is from the distribution of normal features. 
While these normality-centered approaches have demonstrated effectiveness in capturing normality, they share a fundamental limitation: neither \textbf{explicitly learns a defect criterion that characterizes anomalies as relation-level inconsistencies with respect to surrounding structures}.
As illustrated in Figure~\ref{fig:figure1}(a), such ambiguity leads to two typical failure modes: rare but valid normal regions exceed the decision threshold (false positives), while true defects remain within the normal boundary and are missed (false negatives).
These failure modes indicate that deviation from the normal distribution alone is insufficient for reliable anomaly detection — without directional guidance, any such deviation may be mistaken for a true defect.
This limitation becomes more pronounced in practical scenarios, where most existing methods are trained in a category-specific manner~\cite{zhou2024r3d, liang2025taming,  ye2025po3ad, liang2025look} — a separate model per object category — making them ill-suited for real-world systems that require \textbf{unified (multi-category)} and \textbf{cross-domain (unseen-category)} generalization. Under such scenarios, increasingly diverse and entangled normal feature distributions further blur decision boundaries, exacerbating both false positives and false negatives.

\begin{wrapfigure}{r}{0.6\textwidth}
    \centering
    \vspace{-0.4cm}
    \includegraphics[width=0.6\textwidth]{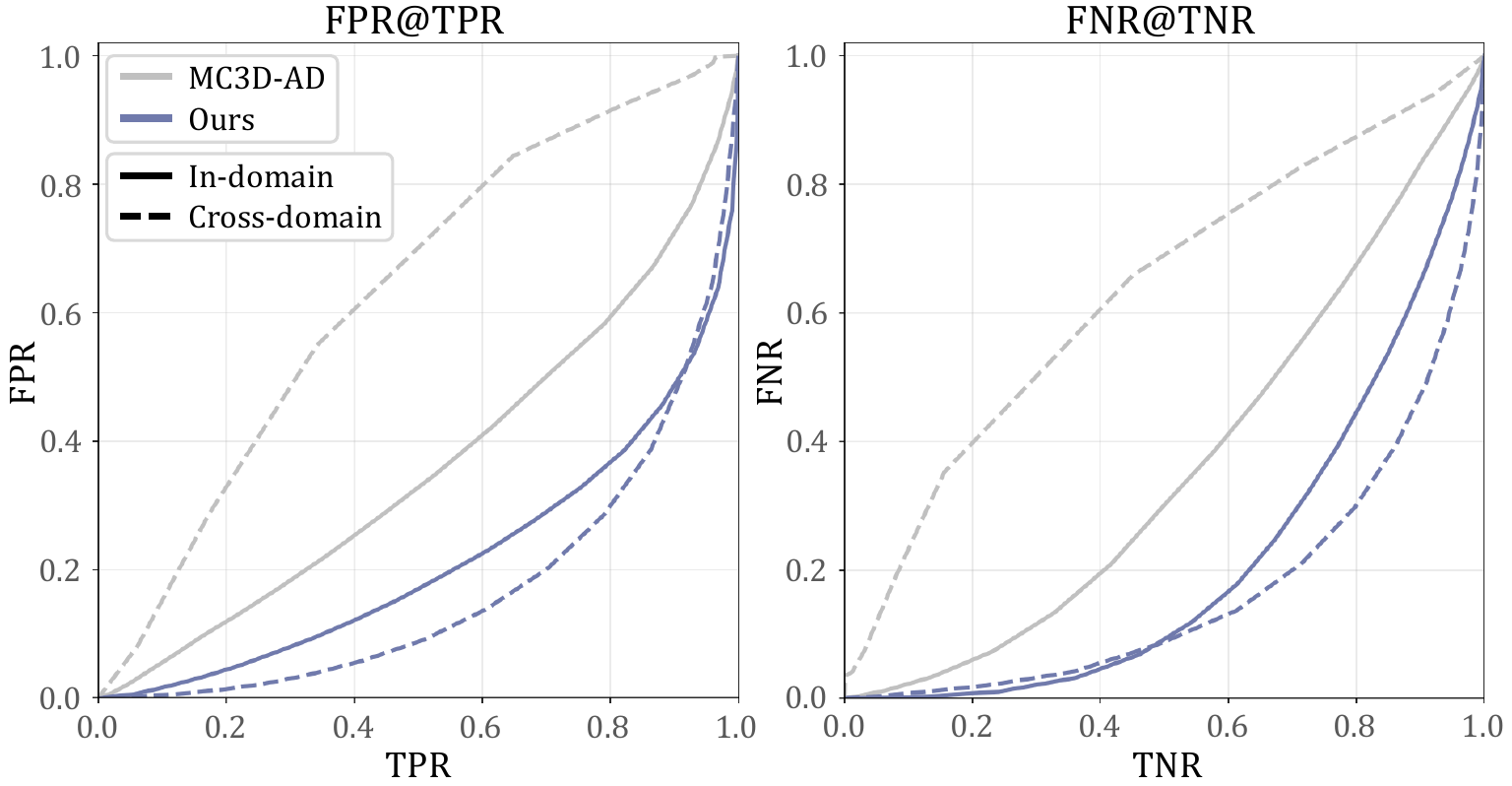}
    \vspace{-0.4cm}
    \caption{Comparison of False Positive Rate (FPR) and False Negative Rate (FNR) between two unified 3DAD models, MC3D-AD and our method.}
    \vspace{-0.1cm}
    \label{fig:figure2}
\end{wrapfigure}

We analyze these challenges in Figure~\ref{fig:figure2} using MC3D-AD~\cite{ijcai2025p94}, a representative unified 3DAD model.
The figure reports FPR at a fixed true positive rate (TPR) and FNR at a fixed true negative rate (TNR) to quantify false alarms on normal regions and missed detections of anomalies, respectively.
As shown, MC3D-AD exhibits high FPR and FNR, indicating that normal regions are frequently misclassified as anomalies while true defects are often missed. 
This degradation is further amplified in cross-domain settings, as MC3D-AD learns category-dependent normality without explicitly learning a defect criterion based on relational inconsistencies, limiting its generalization to unseen domains.


Motivated by this perspective, our work aims to learn category-agnostic defect cues rather than defining diverse defect patterns individually. We argue that many industrial 3D defects are revealed less by the absolute abnormality of a local patch than by inconsistencies that break the geometric relations with surrounding structures. We utilize pseudo-anomalies for supervision, using controlled violations of inter-structural relations to train the model to distinguish structurally coherent regions from relation-violating ones, rather than to simulate the full diversity of possible defect appearances. This formulation provides the model with an explicit criterion for defect identification grounded in \textbf{relational inconsistency}, rather than relying solely on deviation from a learned normal distribution.

To realize this idea, we propose \textbf{GRIM}, which incorporates two key designs to effectively capture the relational nature of defects.
First, we represent each 3D sample as a relational graph composed of local structures and refine regional features via \textbf{Edge-aware Graph Refinement (EGR)}, which performs relation-aware message passing to encode geometric relationships among neighboring regions. 
This design enables the model to capture not only the absolute representation of each local region, but also the structural compatibility reflected in their relative arrangement and connectivity.
Second, we introduce \textbf{Cluster-Deviation Modeling (CDM)}, which measures each region's anomaly score relative to clusters of structurally similar regions, rather than against the entire normal distribution. 
By clustering regional features and quantifying their deviation from cluster centers, CDM captures distribution-level inconsistencies at a finer granularity, identifying regions that are relationally incompatible within their structural peer group even when their absolute feature values appear normal. 
As a result, Figure~\ref{fig:figure1}(b) shows that our method learns an explicit defect criterion, leading to a more discriminative decision boundary and point-level anomaly scores that better separate normal and anomalous regions, thereby reducing both false positives and false negatives.

Extensive experiments validate the effectiveness of our framework, showing consistent improvements over prior state-of-the-art methods with gains of \textbf{7.4}/\textbf{4.7} and \textbf{7.0}/\textbf{3.5} in object-/point-level AUROC on Anomaly-ShapeNet and Real3D-AD, respectively. 
Furthermore, under cross-domain setting, our method achieves \textbf{83.6}/\textbf{89.6} when trained on Anomaly-ShapeNet and evaluated on Real3D-AD, and \textbf{91.7}/\textbf{78.9} in the reverse setting, demonstrating strong generalization in 3D anomaly detection.

Our contributions are summarized as follows:
\begin{itemize}
\item We revisit 3D anomaly detection by defining defects as relation-level inconsistencies, and show that normality-centered approaches lead to ambiguous decision boundaries, resulting in increased false positives and negatives, especially in unified and cross-domain settings.
 \item We propose a \textbf{relational inconsistency} modeling framework that learns an explicit, category-agnostic defect criterion via pseudo-anomalies, instantiated by two key modules: \textbf{Edge-aware Graph Refinement (EGR)} for relation-aware feature learning and \textbf{Cluster-Deviation Modeling (CDM)} for capturing relative distributional inconsistencies.
\item We validate our approach through extensive experiments on Real3D-AD and Anomaly-ShapeNet, achieving consistent gains of \textbf{7.0}/\textbf{3.5} and \textbf{7.4}/\textbf{4.7} in object-/point-level AUROC over prior state-of-the-art methods, respectively. Our method further demonstrates strong generalization in 3D anomaly detection, particularly under cross-domain settings.
\end{itemize}
\section{Related Work}
\subsection{Point Cloud 3D Anomaly Detection}
While 2D anomaly detection has been extensively studied~\cite{he2024mambaad, yao2024resad, guo2025dinomaly, wei2025uninet, fan2025salvaging, you2022unified, zhao2023omnial, lu2023hierarchical}, recent efforts have extended these ideas to 3D data to overcome blind spots in 2D representations.
With the emergence of high-resolution 3D inspection benchmarks~\cite{liu2023real3d, li2024towards, bergmann2021mvtec, bonfiglioli2022eyecandies}, point-cloud 3D anomaly detection (3DAD) has gained increasing attention. Existing methods can be broadly categorized into feature-similarity and reconstruction-based approaches. 
Feature-similarity methods~\cite{liu2023real3d, wang2023multimodal, zhu2024towards, liang2025look, bergmann2023anomaly, horwitz2023back, chu2023shape, liu2025template3d} estimate normality by comparing test features with stored normal representations, but their reliance on memory banks limits scalability to unified settings.
Reconstruction-based methods~\cite{li2024towards, zhou2024r3d, ijcai2025p94, liang2025taming} instead learn to recover normal geometry or features, using reconstruction error as anomaly scores; however, they often suffer from over-generalization, reconstructing defects while over-penalizing rare normal structures. 
Despite these diverse designs, most existing approaches identify anomalies as deviations from learned normality rather than explicitly modeling what constitutes a defect. 
This limitation becomes critical in unified and cross-domain settings, where diverse normal patterns blur decision boundaries, leading to misclassification of both normal variations and true defects. 
In contrast, our method learns an explicit, category-agnostic defect criterion based on relation-level inconsistencies.

\subsection{Learning with Synthetic Anomalies}
Synthetic anomalies are widely used to address the scarcity of real defect annotations. 
In 2D anomaly detection, prior work generates image- or feature-level perturbations to learn discriminative representations~\cite{li2021cutpaste, zavrtanik2021draem, liu2023simplenet, schluter2022natural, zhang2023destseg, zhang2024realnet, hu2024anomalydiffusion}. 
This idea has been extended to 3DAD by synthesizing pseudo-defects in point clouds. For example, R3D-AD~\cite{zhou2024r3d} generates diverse defect shapes for reconstruction-based learning, while PO3AD~\cite{ye2025po3ad} generates normal-vector-guided pseudo-anomaly points and learns point offsets toward normal geometry. 
Other methods employ geometric perturbations such as deformation or stretching to enhance representation learning~\cite{cheng2025boosting, hoang2026vote3d}. 
While effective, these approaches typically treat synthetic anomalies as proxies that mimic real defects. 
In contrast, our work uses pseudo-anomalies not to imitate specific defect patterns, but to provide controlled supervision for violations of geometric relations among normal structures. 
This allows the model to learn a category-agnostic defect criterion grounded in relational inconsistency.
\section{Methods}

\subsection{Problem Statement}
Let $\mathcal{O} = \{1, \ldots, O\}$ denote the set of object categories. For each category $o \in \mathcal{O}$, the training set $\mathcal{D}^{o}_{\mathrm{train}}$ consists solely of unlabeled normal point clouds $\mathcal{P} \in \mathbb{R}^{N \times 3}$, where $N$ denotes the number of points, following the standard anomaly detection protocol. The test set $\mathcal{D}^{o}_{\mathrm{test}}$ contains both normal and anomalous samples, each annotated with an object-level binary label and point-level anomaly labels for localization. In the unified multi-category setting, a single model is trained and evaluated across all categories without category distinction, where $\mathcal{D}_{\mathrm{train}} = \bigcup_{o=1}^{O} \mathcal{D}^{o}_{\mathrm{train}}$ and $\mathcal{D}_{\mathrm{test}} = \bigcup_{o=1}^{O} \mathcal{D}^{o}_{\mathrm{test}}$ aggregate samples from all categories. Given a test point cloud, the model predicts anomaly scores over its constituent point regions for both object-level detection and point-level localization.

\subsection{Pseudo-Anomaly Generation}
\label{sec:pseudo}
To provide explicit supervision for learning defect criteria, we generate pseudo-anomalies from normal point clouds. The generated anomalies serve as controlled local perturbations designed to violate geometric relations among neighboring structures, encouraging the model to learn category-agnostic cues of relation-level inconsistency. Given a normal point cloud $\mathcal{P}$, the pseudo-anomaly generator $\mathcal{A}(\cdot)$ produces a corrupted point cloud $\mathcal{P}^{a}$ along with a point-level binary mask $\mathcal{M} = \{m_n\}_{n=1}^{N}$, where each $m_n$ indicates whether point $\mathbf{p}_n$ has been synthetically modified, serving as point-level supervision during training. To synthesize a localized perturbation, we randomly sample a center point and construct a local patch via its nearest neighbors, confining the deformation while preserving surrounding structures.

We consider three geometric perturbation types: \emph{bulge}, \emph{sink}, and \emph{hole}. Bulge and sink displace the patch along the estimated surface normal in opposite directions, producing protrusion- and depression-like distortions, respectively. Hole moves the central region toward its boundary and pushes it slightly inward while mildly expanding the boundary, forming a cavity-like deformation. These types provide complementary supervision through outward deformation, inward deformation, and local surface-support depletion. Their purpose is to expose violations of local geometric consistency, not to directly approximate the full distribution of real defects or constitute a complete defect taxonomy. Implementation details and visual examples are provided in Section~\ref{pseudo}, while their effectiveness as training supervision is analyzed in Section~\ref{sec:discussion}.

\subsection{Edge-aware Graph Refinement}
\begin{figure}[t]
    \centering
    \includegraphics[width=1\linewidth]{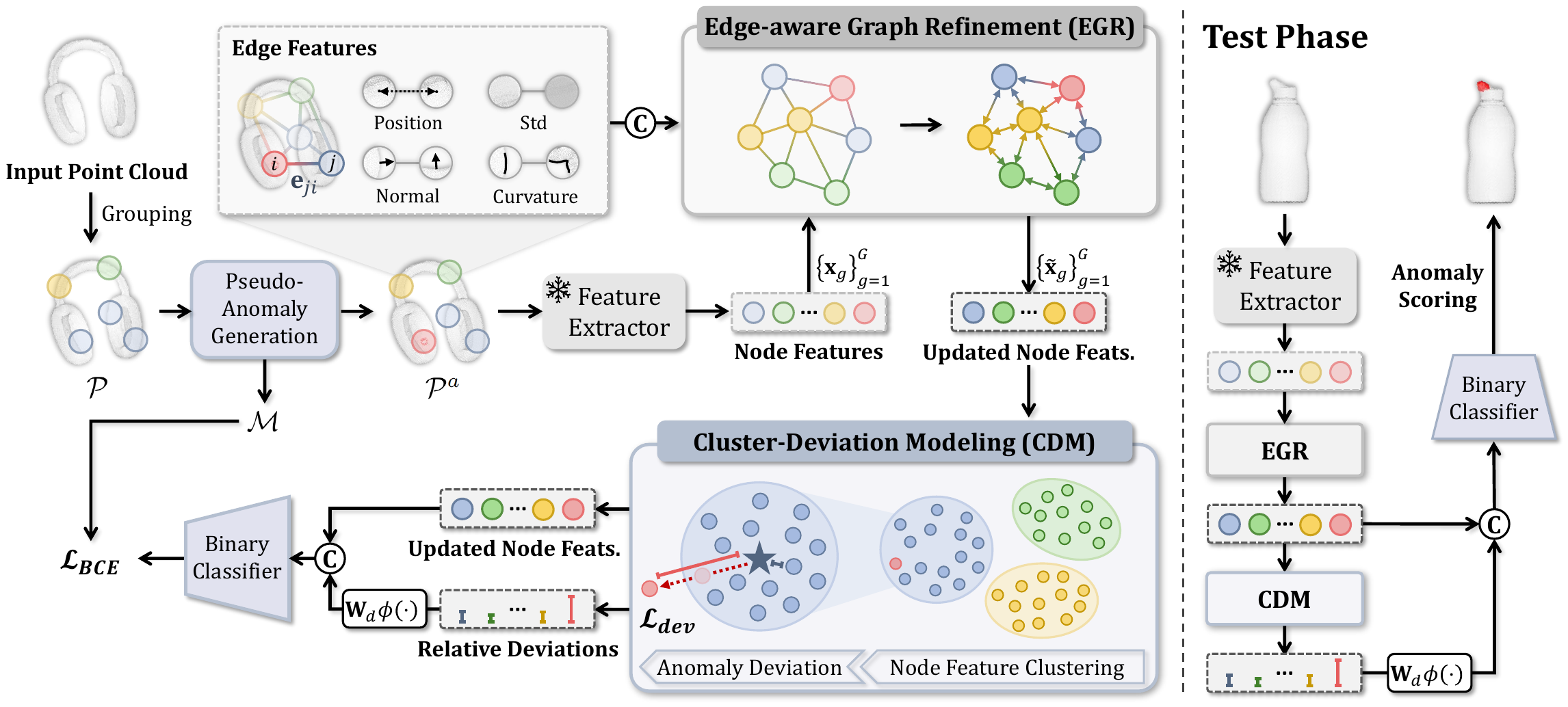}
    \caption{Overview of the proposed method. Given grouped features from an input sample as nodes, we construct edge features encoding geometric relationships between local groups, and perform Edge-aware Graph Refinement (EGR) to obtain refined node features. The refined features are then passed through Cluster-Deviation Modeling (CDM), which captures relative deviations within clusters, and fed into the classifier together with the features.}
    \label{fig:figure3}
\end{figure}

Detecting anomalies in 3D point clouds requires not only characterizing individual local regions, but also understanding their geometric relationships with neighboring structures. To this end, we propose \textbf{Edge-aware Graph Refinement (EGR)}, illustrated in Figure~\ref{fig:figure3}, which models each point cloud as a relational graph of local regions and performs relation-aware message passing to encode structural compatibility among neighboring nodes.

\noindent{\textbf{Node Features.}}
Given an input point cloud $\mathcal{P}$ (or $\mathcal{P}^{a}$ in the case of pseudo-anomaly samples), we first partition it into $G$ local groups $\{\mathcal{P}^{g}\}_{g=1}^{G}$ by selecting representative centers via Farthest-Point Sampling (FPS)~\cite{qi2017pointnet++} and collecting the $k$ nearest neighbors around each center. For each group, the point coordinates are normalized by subtracting the corresponding sampling center, yielding a locally translated neighborhood $\widetilde{\mathcal{P}}^{g} = \{\mathbf{p} - \mathbf{c}_g \mid \mathbf{p} \in \mathcal{P}^{g}\}$. Each normalized group is then encoded using a pretrained local encoder $E(\cdot)$ (\emph{e.g.}, PointMAE~\cite{pang2023masked}), producing a group-level feature $\mathbf{x}_g = E(\widetilde{\mathcal{P}}^{g}) \in \mathbb{R}^{D}$, where $D$ denotes the feature dimension. The resulting features $\{\mathbf{x}_g\}_{g=1}^{G}$ are subsequently adopted as node features in the graph module, establishing a strict one-to-one correspondence between local point groups and graph nodes.

\noindent{\textbf{Edge Features.}}
For each group $\mathcal{P}^{g}$, we compute a geometric descriptor comprising the group centroid $\boldsymbol{\mu}_g$, the coordinate-wise standard deviation $\boldsymbol{\sigma}_g$, the surface normal $\mathbf{n}_g$, and the curvature $\kappa_g$. The surface normal and curvature are estimated via eigendecomposition of the covariance matrix of the grouped points: the normal $\mathbf{n}_g$ is taken as the eigenvector associated with the smallest eigenvalue, and the curvature $\kappa_g$ is defined as the ratio of the smallest eigenvalue to the sum of all eigenvalues. For a directed group pair $(j, i)$, the pairwise relationship feature is defined as
\begin{equation}
    \mathbf{e}_{ji} = \left[\, (\boldsymbol{\mu}_i - \boldsymbol{\mu}_j) \,\Vert\, (\boldsymbol{\sigma}_i - \boldsymbol{\sigma}_j) \,\Vert\, (1 - \left| \mathbf{n}_i^\top \mathbf{n}_j \right|) \,\Vert\, \left| \kappa_i - \kappa_j \right| \,\right],
\end{equation}
where $[\,\cdot\,\Vert\,\cdot\,]$ denotes feature-wise concatenation. Specifically, $\boldsymbol{\mu}_i - \boldsymbol{\mu}_j$ encodes the relative spatial displacement between the two groups; $\boldsymbol{\sigma}_i - \boldsymbol{\sigma}_j$ characterizes the discrepancy in local point spread; $1 - |\mathbf{n}_i^\top \mathbf{n}_j|$ quantifies the degree of normal inconsistency in a sign-invariant manner; and $|\kappa_i - \kappa_j|$ reflects the variation in local surface curvature. The resulting pairwise feature $\mathbf{e}_{ji}$ serves as the edge feature in the subsequent graph module.

\noindent{\textbf{Edge-aware Message Passing.}}
The graph nodes correspond to the group features $\{\mathbf{x}_g\}_{g=1}^{G}$, and edges are constructed by connecting each node to its $K$ nearest neighbors among the group centroids, forming a directed $K$-NN graph. We denote by $\mathcal{N}(i)$ the set of source nodes connected to node $i$. Let $\mathbf{x}_i^{(\ell)} \in \mathbb{R}^{D}$ denote the node feature of group $i$ at layer $\ell$, and let $\mathbf{e}_{ji}^{(\ell)}$ denote the edge feature encoding the pairwise relationship between groups $j$ and $i$. At each layer, the edge representation is first refined by aggregating the features of both endpoint nodes alongside the current edge feature:
\begin{equation}
    \tilde{\mathbf{e}}_{ji}^{(\ell)} = \mathrm{MLP}_e\!\left(\left[\, \mathbf{x}_i^{(\ell)} \,\Vert\, \mathbf{e}_{ji}^{(\ell)} \,\Vert\, \mathbf{x}_j^{(\ell)} \,\right]\right).
\end{equation}
Concurrently, an attention coefficient for the directed edge $j \rightarrow i$ is computed from the destination-node query and the corresponding edge feature:
\begin{equation}
    \alpha_{ji}^{(\ell)} = \operatorname{softmax}_{j \in \mathcal{N}(i)} \phi\!\left(\left[\, \mathrm{MLP}_q\!\left(\mathbf{x}_i^{(\ell)}\right) \,\Vert\, \mathrm{MLP}_r\!\left(\tilde{\mathbf{e}}_{ji}^{(\ell)}\right) \,\right]\right),
\end{equation}
where $\mathrm{MLP}_q(\cdot)$ and $\mathrm{MLP}_r(\cdot)$ are learnable projections, $\phi(\cdot)$ is a learnable scoring function, and the softmax is normalized over all incoming neighbors $\mathcal{N}(i)$. The aggregated neighborhood signal at node $i$ is then obtained by summing attention-weighted messages from all incoming neighbors:
\begin{equation}
    \mathbf{a}_i^{(\ell)} = \sum_{j \in \mathcal{N}(i)} \alpha_{ji}^{(\ell)} \odot \mathrm{MLP}_v\!\left(\mathbf{x}_j^{(\ell)}\right),
\end{equation}
where $\mathrm{MLP}_v(\cdot)$ is a learnable projection.
The node feature is subsequently updated via a residual connection, $\mathbf{x}_i^{(\ell+1)} = \mathbf{x}_i^{(\ell)} + \mathrm{MLP}_n\!\left(\left[\, \mathbf{x}_i^{(\ell)} \,\Vert\, \mathbf{a}_i^{(\ell)} \,\right]\right)$, and the edge feature is similarly refined as $\mathbf{e}_{ji}^{(\ell+1)} = \mathbf{e}_{ji}^{(\ell)} + \tilde{\mathbf{e}}_{ji}^{(\ell)}$. Through this iterative process, the graph module progressively refines group-level representations by propagating relation-conditioned messages along directed edges and accumulating neighborhood information at each node, producing the final refined node features $\tilde{\mathbf{x}}_g$.

\subsection{Cluster-Deviation Modeling}
While \textbf{EGR} captures geometric relationships among neighboring regions, detecting anomalies also requires understanding whether a region is consistent within the broader distribution of structurally similar regions. To this end, we propose \textbf{Cluster-Deviation Modeling (CDM)}, as shown in Figure~\ref{fig:figure3}, which clusters regional features and quantifies each region's deviation from its cluster center. This allows the model to identify anomalies as regions that are relationally incompatible within a set of structurally similar normal structures, rather than relying solely on absolute feature values.

\noindent{\textbf{Node Feature Clustering.}}
Let $\hat{\mathbf{x}}_g = \tilde{\mathbf{x}}_g / \lVert \tilde{\mathbf{x}}_g \rVert_{2}$ denote the $\ell_2$-normalized node feature obtained after graph refinement. We perform clustering on $\{\hat{\mathbf{x}}_g\}_{g=1}^{G}$ into $R$ clusters, where each cluster groups nodes with structurally similar feature representations. The cluster assignment of node $g$ is denoted as $c(g) \in \{1, \dots, R\}$, and the node set of the $r$-th cluster is defined as $\mathcal{C}_r = \{g \mid c(g) = r\}$. The center of each cluster is computed as the $\ell_2$-normalized mean of its member node features:
\begin{equation}
    \mathbf{z}_r = \operatorname{Norm}\!\left( \frac{1}{|\mathcal{C}_r|} \sum_{g \in \mathcal{C}_r} \hat{\mathbf{x}}_g \right),
\end{equation}
where $\operatorname{Norm}(\cdot)$ denotes $\ell_2$ normalization. The cluster center $\mathbf{z}_r$ serves as a reference representation for structurally similar regions within that cluster. Each node is then assigned the center of its corresponding cluster, $\bar{\mathbf{z}}_g = \mathbf{z}_r$ where $c(g) = r$, which is used in the subsequent deviation modeling. The same clustering procedure is applied to each test sample at inference time.

\noindent{\textbf{Distance-based Deviation Modeling}.}
Given the assigned cluster center $\bar{\mathbf{z}}_g$, we quantify the deviation of each node from its cluster center via the cosine distance:
\begin{equation}
    d_g = 1 - \hat{\mathbf{x}}_g^\top \bar{\mathbf{z}}_g.
\end{equation}
The deviation $d_g$ is then normalized within each cluster to produce a cluster-relative score, capturing how much each node deviates from the representative structure of its group. This relative deviation is more informative than absolute feature values, as anomalous regions are expected to exhibit greater inconsistency with respect to the surrounding normal structures. The scalar $d_g$ is subsequently transformed into a sinusoidal positional embedding $\phi(d_g)$, which maps the scalar deviation into a high-dimensional representation to capture subtle deviation differences more expressively, and projected via a learnable linear map $\mathbf{W}_d$. The resulting deviation embedding is concatenated with the node feature to form the final node representation:
\begin{equation}
    \mathbf{h}_g = \left[\, \tilde{\mathbf{x}}_g \,\Vert\, \mathbf{W}_d \phi(d_g) \,\right],
\end{equation}
which is passed to a shared binary classifier to produce the anomaly score. To ensure that anomalous nodes produce deviation scores that are distinguishable from those of normal nodes within the same cluster, anomalous nodes are explicitly encouraged to deviate from their assigned cluster centers during training, as described in detail in Section~\ref{sec:loss}.

\subsection{Loss Function}
\label{sec:loss}
Let $m_n \in \{0, 1\}$ denote the point-level anomaly label. We lift this supervision to the group level by defining $y_g = \max_{p_n \in \mathcal{P}^{g}} m_n$, such that a group is labeled anomalous if it contains at least one anomalous point. The classification loss is defined in a cluster-balanced manner to prevent dominant clusters from biasing the training signal:
\begin{equation}
    \mathcal{L}_{\mathrm{BCE}} = \frac{1}{|\mathcal{R}|} \sum_{r \in \mathcal{R}} \frac{1}{|\mathcal{C}_r|} \sum_{g \in \mathcal{C}_r} \operatorname{BCE}(\ell_g,\, y_g),
\end{equation}
where $\mathcal{R}$ denotes the set of all clusters and $\operatorname{BCE}(\cdot, \cdot)$ denotes binary cross-entropy applied to logits.

To encourage anomalous nodes to deviate from their assigned cluster centers, we additionally employ a deviation loss. Letting $u_g = \hat{\mathbf{x}}_g^\top \bar{\mathbf{z}}_g$ denote the cosine similarity between node $g$ and its cluster center, the deviation loss penalizes anomalous nodes that remain overly close to their center:
\begin{equation}
    \mathcal{L}_{\mathrm{dev}} = \frac{1}{|\mathcal{G}_1|} \sum_{g \in \mathcal{G}_1} \max(u_g - \delta,\; 0),
\end{equation}
where $\mathcal{G}_1 = \{g \mid y_g = 1\}$ denotes the set of anomalous nodes and $\delta$ is a predefined margin. The overall training objective is a weighted sum of the two losses, $\mathcal{L} = \lambda_{\mathrm{bce}}\, \mathcal{L}_{\mathrm{BCE}} + \lambda_{\mathrm{dev}}\, \mathcal{L}_{\mathrm{dev}}$, where $\lambda_{\mathrm{bce}}$ and $\lambda_{\mathrm{dev}}$ are scalar weighting coefficients.
\begin{table*}[t]
\caption{Quantitative comparison of mean AUROC (\%) at both the object level and point level on Anomaly-ShapeNet and Real3D-AD. The best and second-best results are highlighted in \textbf{bold} and \underline{underline}, respectively. Per-category results are provided in the supplementary material.}
\resizebox{\textwidth}{!}{
\begin{tabular}{c|l|cc|cc}
\toprule
\multicolumn{2}{c|}{\multirow{2.5}{*}{Method}} & \multicolumn{2}{c|}{Anomaly-ShapeNet} & \multicolumn{2}{c}{Real3D-AD} \\
\cmidrule(lr){3-4} \cmidrule(lr){5-6}
\multicolumn{2}{c|}{} & O-AUROC($\uparrow$) & P-AUROC($\uparrow$) & O-AUROC($\uparrow$) & P-AUROC($\uparrow$) \\
\midrule
\multirow{14}{*}{\shortstack{Category-\\Specific}}
& BTF(Raw)~\pub{CVPR'23}~\cite{horwitz2023back}          & 49.3 & 55.0 & 63.5 & 57.1 \\
& BTF(FPFH)~\cite{horwitz2023back}                        & 52.8 & 62.8 & 60.3 & 73.3 \\
& M3DM~\pub{CVPR'23}~\cite{wang2023multimodal}             & 55.2 & 61.6 & 59.4 & 62.0 \\
& PatchCore(FPFH)~\pub{CVPR'22}~\cite{roth2022towards}    & 56.8 & 58.0 & 59.3 & 68.2 \\
& PatchCore(PointMAE)~\cite{roth2022towards}               & 56.2 & 57.7 & 59.4 & 62.0 \\
& CPMF~\pub{PR'24}~\cite{cao2024complementary}               & 55.9 & 57.3 & 58.6 & 75.8 \\
& IMRNet~\pub{CVPR'24}~\cite{li2024towards}         & 66.1 & 65.0 & 72.5 & - \\
& Reg3D-AD~\pub{NeurIPS'23}~\cite{liu2023real3d}     & 57.2 & 66.8 & 70.4 & 70.5 \\
& Group3AD~\pub{ACM MM'24}~\cite{zhu2024towards}    & - & - & 75.1 & 73.5 \\
& R3D-AD~\pub{ECCV'24}~\cite{zhou2024r3d}          & 74.9 & - & 73.4 & - \\
& ISMP~\pub{AAAI'25}~\cite{liang2025look}             & - & 69.1 & 76.7 & 83.6 \\
& PO3AD~\pub{CVPR'25}~\cite{ye2025po3ad}           & 83.9 & \underline{89.8} & 76.5 & 65.0 \\
& PASDF~\pub{ICCV'25}~\cite{zheng2025bridging}           & \underline{90.0} & 89.7 & \underline{80.2} & 74.5 \\
& Reg2Inv~\pub{NeurIPS'25}~\cite{yu2025registration}    & 86.1 & 88.2 & 78.0 & \underline{87.8} \\
\midrule[1.2pt]
\multirow{2}{*}{\textbf{Unified}}
& MC3D-AD~\pub{IJCAI'25}~\cite{ijcai2025p94}       & 84.2 & 75.9 & 78.2 & 76.8 \\
& \cellcolor{pubcolor!25}\textbf{GRIM (Ours)} & \cellcolor{pubcolor!25}\textbf{97.4} & \cellcolor{pubcolor!25}\textbf{94.5} & \cellcolor{pubcolor!25}\textbf{87.2} & \cellcolor{pubcolor!25}\textbf{91.3} \\
\bottomrule
\end{tabular}
}
\label{tab:table1}
\end{table*}

{\setlength{\tabcolsep}{10pt}
\begin{table}[t]
\caption{Quantitative comparison of mean AUROC (\%) under the cross-domain setting. In each $\rightarrow$ pair, the left denotes the training set and the right denotes the test set. Values in \textcolor{darkgray}{( \ )} indicate in-domain performance, and $\Delta$ denotes the performance gap between in-domain and cross-domain results.}
\resizebox{\columnwidth}{!}{
\begin{tabular}{c|cc|cc|cc|cc}
\toprule
\multirow{2.5}{*}{Method} 
& \multicolumn{4}{c|}{Anomaly-ShapeNet $\rightarrow$ Real3D-AD} 
& \multicolumn{4}{c}{Real3D-AD $\rightarrow$ Anomaly-ShapeNet} \\
\cmidrule(lr){2-5} \cmidrule(lr){6-9}
& O-AUROC($\uparrow$) & $\Delta$($\downarrow$) & P-AUROC($\uparrow$) & $\Delta$($\downarrow$)
& O-AUROC($\uparrow$) & $\Delta$($\downarrow$) & P-AUROC($\uparrow$) & $\Delta$($\downarrow$) \\
\midrule
MC3D-AD
& 55.4 {\small\textcolor{darkgray}{(78.2)}} & 22.8 
& 38.3 {\small\textcolor{darkgray}{(76.8)}} & 38.5
& 78.3 {\small\textcolor{darkgray}{(84.2)}} & 5.9
& 48.8 {\small\textcolor{darkgray}{(75.9)}} & 27.1 \\
\rowcolor{pubcolor!25}
\textbf{GRIM (Ours)}
& \textbf{83.6} {\small\textcolor{darkgray}{(87.2)}} & \textbf{3.6}
& \textbf{89.6} {\small\textcolor{darkgray}{(91.3)}} & \textbf{1.7}
& \textbf{91.7} {\small\textcolor{darkgray}{(97.4)}} & \textbf{5.7}
& \textbf{78.9} {\small\textcolor{darkgray}{(94.5)}} & \textbf{15.6} \\
\bottomrule
\end{tabular}
}
\label{tab:table2}
\end{table}}

\section{Experiments}

\subsection{Experiment Settings}
\noindent\textbf{Datasets and Evaluation Metrics.}
We evaluate on two benchmarks: Anomaly-ShapeNet~\cite{li2024towards}, a synthetic dataset with 1,600 samples across 40 categories (4 normal training samples each), and Real3D-AD~\cite{liu2023real3d}, a real-world dataset with 12 categories (4 normal training samples each). Notably, Real3D-AD uses single-view test scans despite 360° training captures, posing a more challenging evaluation setting. Following prior work~\cite{liu2023real3d, li2024towards}, we report the Area Under the Receiver Operating Characteristic Curve (AUROC) at both the object level (O-AUROC) and point level (P-AUROC) to measure detection and localization performance, where higher values indicate better performance.

\noindent\textbf{Baseline Methods.}
We compare our method against a range of state-of-the-art approaches: BTF~\cite{horwitz2023back}, M3DM~\cite{wang2023multimodal}, PatchCore~\cite{roth2022towards}, CPMF~\cite{cao2024complementary}, IMRNet~\cite{li2024towards}, Reg3D-AD~\cite{liu2023real3d}, Group3AD~\cite{zhu2024towards}, R3D-AD~\cite{zhou2024r3d}, ISMP~\cite{liang2025look}, PO3AD~\cite{ye2025po3ad}, PASDF~\cite{zheng2025bridging}, Reg2Inv~\cite{yu2025registration}, and MC3D-AD~\cite{ijcai2025p94}. Among these, MC3D-AD is a unified model trained jointly across all categories, where diverse normal distributions across categories must be handled within a single model, making it a fundamentally more challenging setting than the category-specific paradigm. The remaining methods follow a category-specific paradigm, requiring a separate model to be trained for each object category. Results for all baselines are taken from their respective papers or reproduced using publicly available code.

\subsection{Quantitative Results}
\noindent\textbf{In-Domain.} Table~\ref{tab:table1} reports the mean O-AUROC and P-AUROC across all categories on Anomaly-ShapeNet and Real3D-AD. Despite being trained as a single unified model, our method consistently surpasses all category-specific baselines. On Anomaly-ShapeNet, we achieve gains of \textbf{7.4} in O-AUROC over PASDF~\cite{zheng2025bridging} and \textbf{4.7} in P-AUROC over PO3AD~\cite{ye2025po3ad}, the previous best-performing methods in each metric. The same trend holds on Real3D-AD, where our method outperforms PASDF by \textbf{7.0} in O-AUROC and Reg2Inv~\cite{yu2025registration} by \textbf{3.5} in P-AUROC. Furthermore, while MC3D-AD~\cite{ijcai2025p94} retains the training efficiency of a unified model, it lags behind category-specific methods in detection performance, revealing a trade-off. Our method overcomes this limitation: we outperform MC3D-AD by \textbf{13.2} and \textbf{18.6} in O/P-AUROC on Anomaly-ShapeNet, and by \textbf{8.9} and \textbf{14.4} on Real3D-AD. These results demonstrate that training efficiency and detection accuracy need not be at odds, affirming the potential of unified frameworks for 3D anomaly detection.

\noindent\textbf{Cross-Domain.} Table~\ref{tab:table2} evaluates generalization capability under a cross-domain setting, where the training and test datasets are disjoint. We consider two transfer directions --- training on Anomaly-ShapeNet and evaluating on Real3D-AD without any fine-tuning or domain-specific adaptation, and the reverse --- and compare against the prior unified method MC3D-AD~\cite{ijcai2025p94}. When trained on Anomaly-ShapeNet and evaluated on Real3D-AD, our method outperforms MC3D-AD by \textbf{28.2} in O-AUROC and \textbf{51.3} in P-AUROC. Remarkably, our model even surpasses all baselines in Table~\ref{tab:table1} that are trained directly on Real3D-AD. Furthermore, while MC3D-AD suffers a large performance drop in the cross-domain setting ($\Delta$ O/P-AUROC: 22.8 / 38.5), our method remains highly stable ($\Delta$ O/P-AUROC: \textbf{3.6} / \textbf{1.7}), demonstrating that our method effectively learns category-agnostic defect cues that generalize across domains, consistent with our core claim. The reverse transfer direction exhibits a similar trend, further corroborating this finding.

\begin{figure}[t]
    \centering
    \includegraphics[width=1\linewidth]{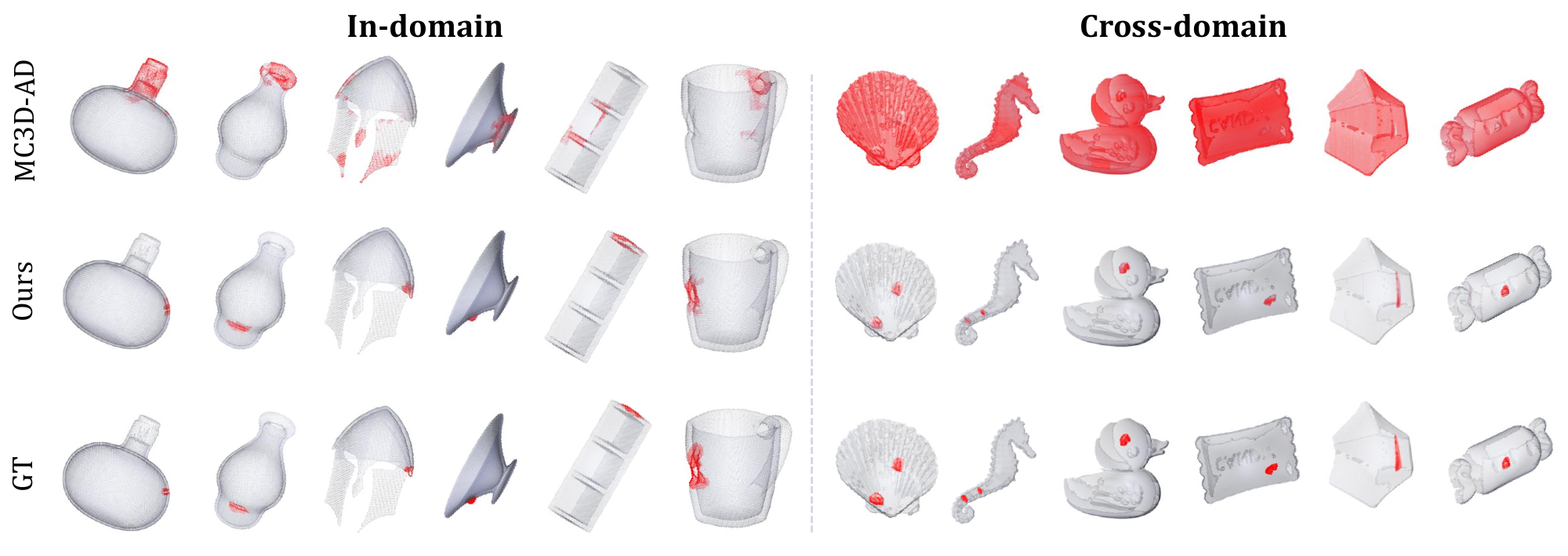}
    \caption{Qualitative comparison across various object categories, showing in-domain results (trained and tested on Anomaly-ShapeNet) and cross-domain results (trained on Anomaly-ShapeNet and tested on Real3D-AD). Red colored regions indicate detected anomalies.}
    \label{fig:figure4}
\end{figure}


\subsection{Qualitative Results}
Figure~\ref{fig:figure4} presents qualitative comparisons between our method and MC3D-AD~\cite{ijcai2025p94} under both in-domain and cross-domain settings. MC3D-AD, as a reconstruction-based approach lacking explicit defect supervision, exhibits two characteristic failure modes: incorrectly identifying rare yet normal structure variations as anomalies, and failing to detect anomalous regions that resemble normal surfaces. These limitations are further exacerbated in the cross-domain setting, where MC3D-AD tends to misclassify the majority of points as anomalous. In contrast, our method consistently produces precise anomaly localization in both settings, validating that our approach successfully learns category-agnostic defect representations that generalize across domain shifts. An extensive set of additional qualitative results is provided in the supplementary material.

\subsection{Discussion}
\label{sec:discussion}

\begin{wrapfigure}{r}{0.48\columnwidth}
    \vspace{-0.5cm}
    \centering
    \includegraphics[width=0.37\columnwidth]{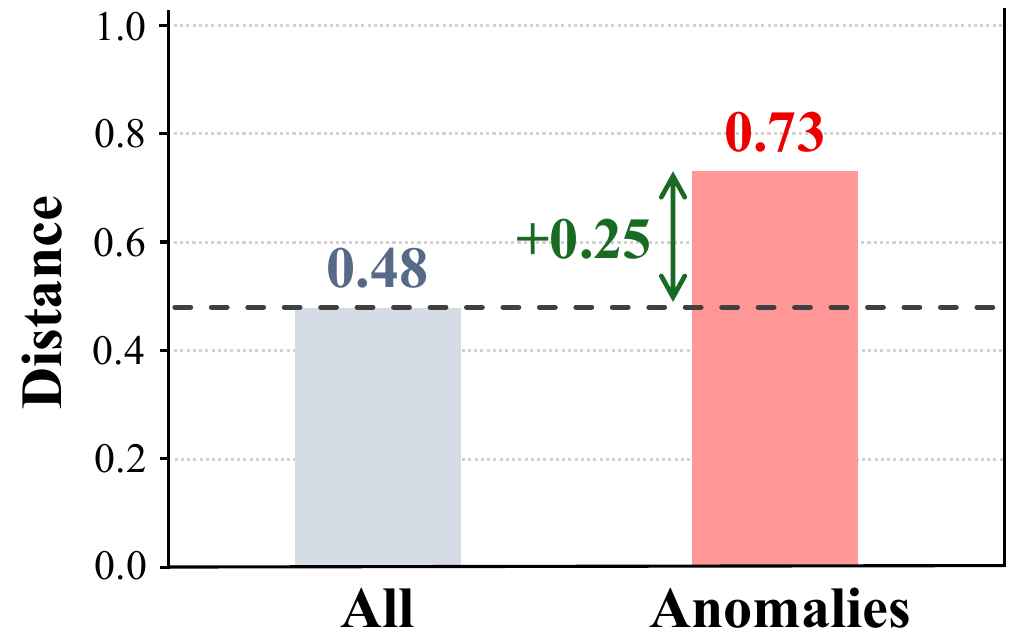}
    \caption{Average distances between features and cluster centers.}
    \vspace{-0.4cm}
    \label{fig:figure5}
\end{wrapfigure}

\textbf{Effect of CDM on Center Distance.} Figure~\ref{fig:figure5} compares the average distance to the cluster center between all features and anomalous features alone, evaluated on the Anomaly-ShapeNet test set. Anomalous features exhibit a notably higher mean distance (0.73) than overall average (0.48), confirming that anomalous features tend to deviate further from their cluster centers under CDM. This deviation serves as a discriminative cue that distinguishes anomalous points from normal ones, validating the design rationale behind CDM.

\begin{wraptable}{r}{0.55\columnwidth}
\vspace{-1.0\intextsep}
\caption{Ablation studies on the proposed components. O-AUROC (O-R) and P-AUROC (P-R) are reported on Anomaly-ShapeNet (A) and Real3D-AD (R).}
\resizebox{0.55\columnwidth}{!}{
\begin{tabular}{cc|cc|cc}
\toprule
\multirow{2.5}{*}{EGR} & \multirow{2.5}{*}{CDM} & \multicolumn{2}{c|}{\small In-Domain (A $\rightarrow$ A)} & \multicolumn{2}{c}{\small Cross-Domain (A $\rightarrow$ R)} \\
\cmidrule(lr){3-4} \cmidrule(lr){5-6}
& & O-R($\uparrow$) & P-R($\uparrow$) & O-R($\uparrow$) & P-R($\uparrow$) \\
\midrule
     &            & 89.5 & 58.1 & 56.9 & 42.8 \\
\checkmark &            & 94.3 & 86.4 & 59.4 & 53.0 \\
     & \checkmark & 96.6 & 79.8 & 62.3 & 64.9 \\
\rowcolor{pubcolor!25}
\checkmark & \checkmark & \textbf{97.4} & \textbf{94.5} & \textbf{83.6} & \textbf{89.6} \\
\bottomrule
\vspace{-0.6cm}
\end{tabular}
}
\label{tab:table3}
\end{wraptable}



\textbf{Analysis of Proposed Methods.} Table~\ref{tab:table3} reports ablation results under both in-domain and cross-domain settings. Our baseline consists solely of a point backbone and a classifier. Adding Edge-aware Graph Refinement (EGR) yields consistent gains across both settings, with a \textbf{28.3} improvement in in-domain P-AUROC, demonstrating the effectiveness of modeling geometric relationships among local regions via graph-based representations. Adding Cluster-Deviation Modeling (CDM) further improves overall performance, with a pronounced gain of \textbf{22.1} in cross-domain P-AUROC, suggesting that exploiting relative deviation within clusters of structurally similar regions is more effective than relying solely on absolute feature representations for cross-domain generalization. When all components are combined, the substantial gains across all metrics indicate that the two modules operate synergistically, contributing to the performance.

\begin{wraptable}{r}{0.6\columnwidth}
\centering
\vspace{-1.3em}
\caption{Ablation studies on anomaly types.}
\resizebox{0.6\columnwidth}{!}{
\begin{tabular}{ccc|cc|cc}
\toprule
\multirow{2}{*}{Bulge} & \multirow{2}{*}{Sink} & \multirow{2}{*}{Hole} & \multicolumn{2}{c|}{\small In-Domain (A $\rightarrow$ A)} & \multicolumn{2}{c}{\small Cross-Domain (A $\rightarrow$ R)} \\
\cmidrule(lr){4-5} \cmidrule(lr){6-7}
& & & O-R($\uparrow$) & P-R($\uparrow$) & O-R($\uparrow$) & P-R($\uparrow$) \\
\midrule
\checkmark &            &            & 96.9 & 94.1 & 77.8 & 88.0 \\
           & \checkmark &            & 98.5 & 95.9 & 81.2 & 81.8 \\
           &            & \checkmark & 95.9 & 93.9 & 78.4 & 86.8 \\
\checkmark & \checkmark &            & 96.9 & 94.1 & 81.8 & 89.0 \\
\checkmark &            & \checkmark & 98.6 & 96.0 & 80.3 & 87.9 \\
           & \checkmark & \checkmark & \textbf{99.2} & \textbf{98.1} & 80.3 & 86.7 \\
\rowcolor{pubcolor!25}
\checkmark & \checkmark & \checkmark & 97.4 & 94.5 & \textbf{83.6} & \textbf{89.6} \\
\bottomrule
\vspace{-0.6cm}
\end{tabular}
}
\label{tab:table4}
\end{wraptable}

\textbf{Analysis of Anomaly Types.} Table~\ref{tab:table4} compares different combinations of pseudo-anomaly types under in-domain and cross-domain settings. Notably, using any single anomaly type alone already yields competitive performance, with in-domain results remaining comparable to the full three-type setting and only a marginal cross-domain decline. Also, certain two-type combinations outperform the three-type configuration in-domain. Thus, increasing anomaly diversity does not produce monotonic improvements across all settings. These observations support the use of complementary perturbations as supervision for learning a transferable relational criterion, without requiring exhaustive simulation of real defect appearances. Here, the empirical sufficiency of the three primitives refers to their effectiveness as training supervision in the evaluated settings, rather than coverage of all real-world defect categories. Additional defect-type breakdowns and cross-type supervision experiments are presented in Section~\ref{discussion}.

\vspace{-0.2cm}
\section{Conclusion}
\vspace{-0.2cm}
We presented \textbf{GRIM}, a relational inconsistency modeling framework for 3D anomaly detection, built on the insight that defects manifest as violations of geometric consistency among neighboring structures rather than mere deviations from normality. Our two proposed modules, EGR and CDM, work synergistically to capture relational inconsistencies at both the structural and cluster levels, enabling the model to learn an explicit, category-agnostic defect criterion through pseudo-anomaly supervision. Experiments demonstrate improvements over prior methods across in- and cross-domain settings, validating relational inconsistency modeling as a promising direction for 3D anomaly detection.

\section*{Acknowledgments}
This work was supported by the National Research Foundation of Korea (NRF) grant funded by the Korea government (MSIT) (RS-2024-00456589, RS-2026-25516683).


\bibliographystyle{plain}
\bibliography{references}


\clearpage
\newpage
\section{Appendix}


\textbf{Contents}
\vspace{-0.1cm}
\begin{itemize}[leftmargin=*, itemsep=2pt]
    \item \textbf{\ref{pseudo}} Pseudo-Anomaly Generation Details
    \item \textbf{\ref{implement}} Implementation Details
    \item \textbf{\ref{discussion}} Additional Discussion
    \item \textbf{\ref{limitation}} Limitations and Future Works
    \item \textbf{\ref{Quantitative}} Additional Quantitative Results
    \item \textbf{\ref{Qualitative}} Additional Qualitative Results
\end{itemize}

\subsection{Pseudo-Anomaly Generation Details} 
\label{pseudo}

\begin{algorithm}[h]
\caption{Pseudo-Anomaly Generation Algorithm}
\label{alg:pseudo_anomaly}
\begin{algorithmic}[1]
\Statex \textbf{Input:} Normal point cloud $\mathcal{P}=\{\mathbf{p}_n\}_{n=1}^{N}$
\Statex \textbf{Output:} Corrupted point cloud $\mathcal{P}^{a}$, anomaly mask $\mathcal{M} = \{m_n\}_{n=1}^{N}$

\State Initialize $\mathcal{P}^{a} \leftarrow \mathcal{P}$ and $\mathcal{M}\leftarrow \mathbf{0}$
\State Randomly sample a seed point from $\mathcal{P}$
\State Construct a local patch $\mathcal{Q}$ using $K$ nearest neighbors of the seed point
\State Estimate the local surface normal $\mathbf{n}$ of $\mathcal{Q}$ by PCA
\State Randomly select anomaly type 
$\tau \in \{\texttt{bulge}, \texttt{sink}, \texttt{hole}\}$

\If{$\tau = \texttt{bulge}$}
    \State Move points in $\mathcal{Q}$ outward along $\mathbf{n}$ with radial falloff
\ElsIf{$\tau = \texttt{sink}$}
    \State Move points in $\mathcal{Q}$ inward along $-\mathbf{n}$ with radial falloff
\ElsIf{$\tau = \texttt{hole}$}
    \State Move central points of $\mathcal{Q}$ toward the boundary
    \State Push central points slightly inward and expand boundary points mildly
\EndIf

\State Update the modified patch in $\mathcal{P}^{a}$
\State Set $m_n=1$ for modified points and $m_n=0$ otherwise
\State \Return $\mathcal{P}^{a}, \mathcal{M}$

\end{algorithmic}
\end{algorithm}

The overall procedure for pseudo-anomaly generation is summarized in Algorithm~\ref{alg:pseudo_anomaly}.
For each normal point cloud, we synthesize pseudo-anomalies by locally perturbing a small neighborhood.
A seed point is randomly selected from $\mathcal{P}$, and its nearest neighbors are used to form a local patch $\mathcal{Q}$. The local perturbation modifies approximately $1\%$ of the points in each object. We estimate the local surface normal $\mathbf{n}$ of $\mathcal{Q}$ using PCA, where the eigenvector corresponding to the smallest eigenvalue is used as the normal direction.
To make the perturbation direction consistent, the normal is oriented outward with respect to the global center of the point cloud.

The pseudo-anomaly generator supports three geometric perturbation primitives: \texttt{bulge}, \texttt{sink}, and \texttt{hole}.
For \texttt{bulge} and \texttt{sink}, each point in the local patch is displaced along the estimated normal direction with a radial falloff:
\begin{equation}
\mathbf{q}_i^{a}
=
\mathbf{q}_i
+
\alpha s w_i \mathbf{n},
\end{equation}
where $\alpha=1$ for \texttt{bulge} and $\alpha=-1$ for \texttt{sink}.
Here, $s \sim \mathcal{U}(0.01,0.03)$ denotes the deformation scale, and the radial falloff weight $w_i$ is defined as
\begin{equation}
w_i =
\left(
1 -
\frac{\|\mathbf{q}_i-\bar{\mathbf{q}}\|_2}
{\max_j \|\mathbf{q}_j-\bar{\mathbf{q}}\|_2 + \epsilon}
\right)^2,
\end{equation}
where $\bar{\mathbf{q}}$ is the patch center.
This falloff makes points near the patch center deform more strongly, while smoothly reducing the displacement toward the patch boundary.

\begin{figure}[t]
    \centering
    \includegraphics[width=1.0\linewidth]{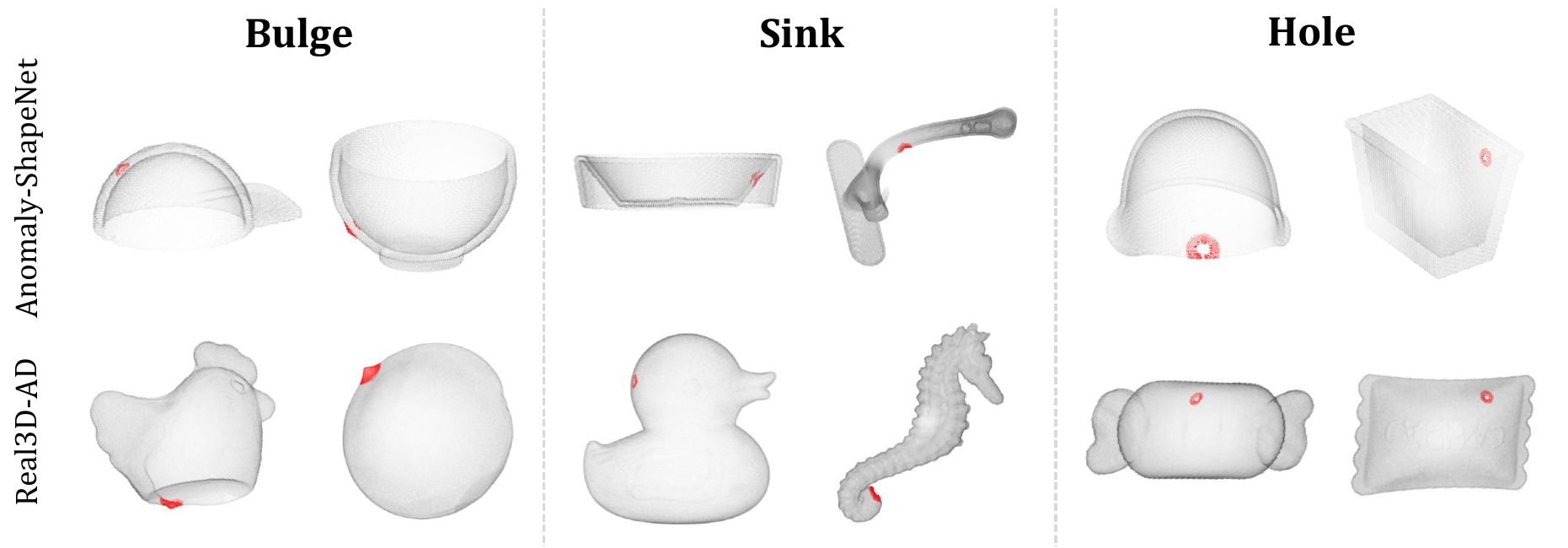}
    \caption{Visualization of pseudo-anomalies generated on Anomaly-ShapeNet and Real3D-AD. Gray points represent object point clouds, and red points denote the perturbed regions used as pseudo-anomaly masks.}
    \label{fig:supp_pseudo}
\end{figure}

For \texttt{hole}, the patch is divided into a central core region and a surrounding boundary ring.
Core points are moved toward their nearest ring points and are additionally pushed inward along $-\mathbf{n}$, while ring points are mildly expanded along their radial directions.
This produces a cavity-like local deformation while preserving the overall locality of the perturbation.
The anomaly mask $\mathcal{M}$ is set to one for synthetically modified points and zero otherwise. Representative examples of the generated pseudo-anomalies are illustrated in Figure~\ref{fig:supp_pseudo}.

After generating the pseudo-anomaly, we further apply a random rotation to the corrupted point cloud.
The same rotation is applied to all points, while the point-level anomaly mask remains unchanged.
This post-generation rotation encourages the model to learn pose-agnostic geometric inconsistency rather than relying on a fixed object orientation.

\subsection{Implementation Details}
\label{implement}

\noindent\textbf{Model setup.}
We adopt PointMAE~\cite{pang2023masked}, pretrained on ModelNet40 (8k), as the local feature extractor. The local feature extractor is initialized from the pretrained checkpoint and kept frozen during training. Each point cloud is partitioned into $G=2048$ groups via FPS~\cite{qi2017pointnet++}, with 128 neighboring points per group. 
For EGR, we construct a 32-NN graph over group centroids and employ 2 graph refinement layers with 4 attention heads and an attention dimension of 384. The edge feature has dimension 8, consisting of centroid difference (dimension 3), standard deviation difference (dimension 3), normal inconsistency (dimension 1), and curvature difference (dimension 1). We use summation aggregation with residual connections.
For CDM, refined node features are $\ell_2$-normalized and clustered into $R=64$ centers using k-means++~\cite{arthur2007k} with 8 iterations. Node-to-center distances are normalized via quantile-based min-max normalization (0.1–0.9 quantiles), then encoded using a sinusoidal embedding with 128 dimensions (temperature 10000, scale 10), followed by a linear projection. The binary classifier is implemented as an MLP with hidden dimensions [256, 128] and dropout 0.1.

\noindent\textbf{Anomaly scoring.}
For the $i$-th test point cloud $P_i = \{\mathbf{p}_{i,n}\}_{n=1}^{N_i}$, the binary classifier predicts a node-level anomaly score $a_{i,g}$ at each sampled group center $\mathbf{c}_{i,g}$, where $g \in \{1,\ldots,G\}$. We first propagate these scores to the original points by assigning each point the score of its nearest sampled node:
\begin{equation}
    g_i^{*}(n)
    =
    \underset{g \in \{1,\ldots,G\}}{\arg\min}
    \left\| \mathbf{p}_{i,n} - \mathbf{c}_{i,g} \right\|_2,
    \qquad
    s_{i,n} = a_{i,g_i^{*}(n)}.
    \label{eq:anomaly_score_propagation}
\end{equation}
We then apply Gaussian smoothing to the propagated score vector $\mathbf{s}_i = [s_{i,1},\ldots,s_{i,N_i}]$ and obtain the object-level anomaly score by taking the maximum of the smoothed point-wise anomaly map:
\begin{equation}
    \widetilde{\mathbf{s}}_i
    =
    \operatorname{GaussianSmooth}_{\sigma}(\mathbf{s}_i),
    \qquad
    S_i^{\mathrm{obj}}
    =
    \max_{1 \leq n \leq N_i} \widetilde{s}_{i,n}.
    \label{eq:object_anomaly_score}
\end{equation}
No sample-wise min--max normalization is applied to the node-level or point-level anomaly scores before computing $S_i^{\mathrm{obj}}$.

For evaluation, min--max normalization is applied only after the object-level scores have been computed, across evaluation samples within each category.
Let $\mathcal{D}_c$ denote the index set of evaluation samples belonging to category $c$. The normalized object-level score is
\begin{equation}
    \widehat{S}_i^{\mathrm{obj}}
    =
    \frac{
        S_i^{\mathrm{obj}}
        -
        \min_{j \in \mathcal{D}_c} S_j^{\mathrm{obj}}
    }{
        \max_{j \in \mathcal{D}_c} S_j^{\mathrm{obj}}
        -
        \min_{j \in \mathcal{D}_c} S_j^{\mathrm{obj}}
    },
    \qquad i \in \mathcal{D}_c.
    \label{eq:category_object_score_normalization}
\end{equation}

\noindent\textbf{Training and evaluation.}
We train the model under the unified setting, where all categories are merged without using category labels. For each normal sample, we generate 4 pseudo-anomalous variants, with bulge, sink, and hole sampled with probabilities 0.4, 0.4, and 0.2, respectively. Random rotation is applied during training.
We use AdamW~\cite{loshchilov2017decoupled} with a learning rate of $1\times10^{-4}$, weight decay $1\times10^{-4}$.
Training is conducted for 100 epochs with a batch size of 1.
The learning rate is decayed using StepLR. The loss weights are set to $\lambda_{\mathrm{bce}}=1.0$ and $\lambda_{\mathrm{dev}}=1.5$. Cross-domain evaluation is conducted by directly applying the trained model to the target domain without fine-tuning or domain-specific adaptation.
All experiments are conducted with PyTorch 1.13.0 and CUDA 11.7 on a single NVIDIA GeForce RTX 3090 GPU, with a training time of approximately 12 hours.

\subsection{Additional Discussion} 
\label{discussion}
\begin{figure}[t]
    \centering
    \includegraphics[width=0.9\linewidth]{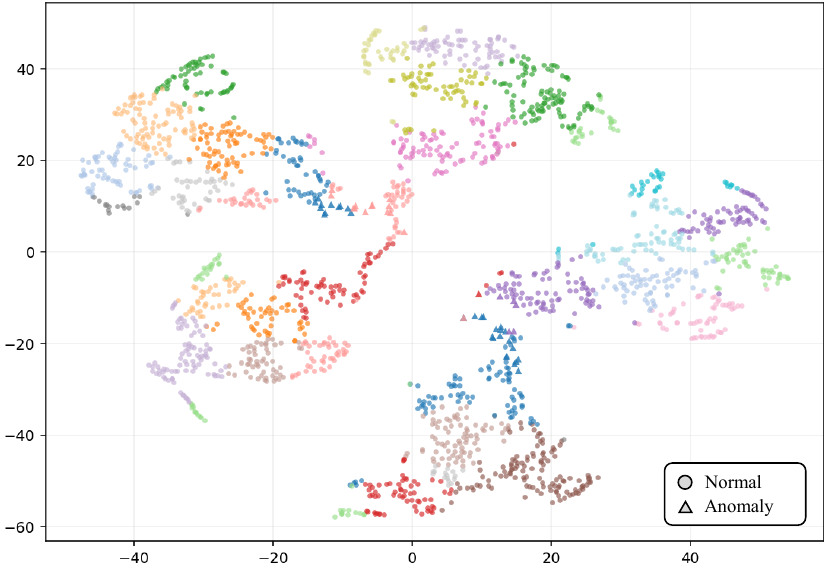}
    \caption{t-SNE visualization of clustering results in CDM. Nodes of the same color belong to the same cluster.}
    \label{fig:supp_cluster}
\end{figure}

\noindent\textbf{Analysis of Clustering in CDM.}
Figure~\ref{fig:supp_cluster} presents a t-SNE visualization of the clustering results in \textbf{Cluster-Deviation Modeling (CDM)}, showing that structurally similar region features are well-separated into distinct clusters. Table~\ref{tab:table5} reports the effect of the number of clusters $R$ on both in-domain and cross-domain performance. In the in-domain setting, performance improves generally as $R$ increases, since finer clustering groups structurally similar regions more precisely, yielding more discriminative deviation scores. A similar trend is observed in the cross-domain setting up to $R=64$.

\begin{table}[t]
\caption{Ablation studies on the number of clusters $R$ on Anomaly-ShapeNet (A) and Real3D-AD (R).}
\centering
\resizebox{0.7\columnwidth}{!}{
\begin{tabular}{c|cc|cc}
\toprule
\multirow{2.5}{*}{$R$} & \multicolumn{2}{c|}{In-Domain (A $\rightarrow$ A)} & \multicolumn{2}{c}{Cross-Domain (A $\rightarrow$ R)} \\
\cmidrule(lr){2-3} \cmidrule(lr){4-5}
& O-AUROC($\uparrow$) & P-AUROC($\uparrow$) & O-AUROC($\uparrow$) & P-AUROC($\uparrow$) \\
\midrule
16  & 96.1 & 89.9 & 63.8 & 73.3 \\
32  & 96.8 & 92.5 & 68.4 & 79.0 \\
\rowcolor{pubcolor!25}
64  & \textbf{97.4} & \textbf{94.5} & \textbf{83.6} & \textbf{89.6} \\
\bottomrule
\end{tabular}
}
\label{tab:table5}
\end{table}

\begin{table}[t]
\caption{Ablation studies on the weighting coefficient $\lambda_{\mathrm{dev}}$.}
\centering
\resizebox{0.7\columnwidth}{!}{
\begin{tabular}{c|cc|cc}
\toprule
\multirow{2.5}{*}{$\lambda_{\mathrm{dev}}$} & \multicolumn{2}{c|}{In-Domain (A $\rightarrow$ A)} & \multicolumn{2}{c}{Cross-Domain (A $\rightarrow$ R)} \\
\cmidrule(lr){2-3} \cmidrule(lr){4-5}
& O-AUROC($\uparrow$) & P-AUROC($\uparrow$) & O-AUROC($\uparrow$) & P-AUROC($\uparrow$) \\
\midrule
0.5 & 96.5 & 93.1 & 81.6 & 88.4 \\
1.0 & 97.3 & 94.5 & 81.8 & 86.7 \\
\rowcolor{pubcolor!25}
1.5 & \textbf{97.4} & \textbf{94.5} & \textbf{83.6} & \textbf{89.6} \\
\bottomrule
\end{tabular}
}
\label{tab:table6}
\end{table}

\noindent\textbf{Analysis of Deviation Weight in CDM.}
Table~\ref{tab:table6} reports the effect of the deviation loss weighting coefficient $\lambda_{\mathrm{dev}}$ on in-domain and cross-domain performance. As $\lambda_{\mathrm{dev}}$ increases, anomalous nodes are more strongly pushed away from their cluster centers, producing deviation scores that are more clearly distinguishable from those of normal nodes, which leads to general in-domain performance gains. In the cross-domain setting, the same tendency holds up to $\lambda_{\mathrm{dev}} = 1.5$; beyond this point, however, an excessively large deviation penalty forces anomalous features to deviate in a manner that is overly tailored to the training distribution, which reduces generalizability under domain shift.

\begin{table}[!ht]
\caption{Comparison of computational complexity.}
\centering
\small
\begin{tabular}{l|l|cccc}
\toprule
\multicolumn{2}{c|}{Method} & AIT $\downarrow$ & FPS $\uparrow$ & \#Params $\downarrow$ & FLOPs $\downarrow$ \\
\midrule
\multirow{3}{*}{\shortstack[l]{Category-\\Specific}}
  & PO3AD   & \textbf{0.31s} & \textbf{3.23} & 433.6M & \textbf{399} \\
  & PASDF   & 0.92s & 1.09 & \textbf{15.9M}  & 624 \\
  & Reg2Inv & 0.78s & 1.28 & 85.3M  & 815 \\
\midrule
\multirow{2}{*}{Unified}
  & MC3D-AD & 0.84s & 1.19 & 34.2M & 417 \\
  & \cellcolor{pubcolor!25}\textbf{GRIM (Ours)}
  & \cellcolor{pubcolor!25}\textbf{0.35s}
  & \cellcolor{pubcolor!25}\textbf{2.86}
  & \cellcolor{pubcolor!25}\textbf{34.0M}
  & \cellcolor{pubcolor!25}\textbf{371} \\
\bottomrule
\end{tabular}
\label{tab:table7}
\end{table}

\noindent\textbf{Computational Complexity.}
Table~\ref{tab:table7} compares average inference time (AIT) and the number of parameters between our method and the category-specific models PO3AD~\cite{ye2025po3ad}, PASDF~\cite{zheng2025bridging}, and Reg2Inv~\cite{yu2025registration}, as well as the previous unified method MC3D-AD~\cite{ijcai2025p94}. AIT is measured as the mean inference time over the 1,312 test samples of Anomaly-ShapeNet on a single NVIDIA RTX 3090 GPU. Our method achieves an AIT of 0.35s, which is more than twice as fast as MC3D-AD, demonstrating that our superior detection performance comes with no computational overhead — in fact, with a clear speed advantage. Furthermore, our model requires fewer parameters than MC3D-AD, highlighting its practical efficiency.

\begin{table}[!ht]
    \centering
    \caption{
        Defect-type breakdown on Anomaly-ShapeNet. 
    }
    \label{tab:pseudo_defect_type_breakdown}
    \small
    \setlength{\tabcolsep}{5pt}
    \begin{tabular}{lcccccc}
        \toprule
        Metric & Bulge & Concavity & Hole & Broken & Bending & Scratch \\
        \midrule
        O-AUROC($\uparrow$)
            & 96.1 & 95.9 & 97.0 & 96.9 & 96.6 & 97.7 \\
        P-AUROC($\uparrow$)
            & 94.2 & 94.6 & 93.3 & 93.8 & 94.8 & 95.3 \\
        \bottomrule
    \end{tabular}
\end{table}

\noindent\textbf{Generalization across Defect Types.} We train the model on Anomaly-ShapeNet using bulge, sink, and hole pseudo-anomalies and evaluate it separately on the six test defect types: bulge, concavity, hole, broken, bending, and scratch. Bulge and hole directly correspond to training primitives, while concavity is geometrically similar to sink and is therefore not treated as an entirely unseen type. Broken, bending, and scratch are not explicitly synthesized by our generator. Table~\ref{tab:pseudo_defect_type_breakdown} reports the results. The model performs well on the defect types not explicitly synthesized during training; in particular, scratch achieves higher O-AUROC and P-AUROC than both bulge and hole. These results provide evidence of transfer across defect types beyond those directly instantiated by the training primitives.

\noindent\textbf{Cross-Type Supervision.} We further evaluate cross-type supervision on Real3D-AD by restricting pseudo-anomaly generation to the hole primitive. The Real3D-AD test set contains bulge and sink defects, but no hole defects. As shown in Table~\ref{tab:pseudo_hole_only_real3d}, hole-only supervision achieves 87.1 O-AUROC and 89.2 P-AUROC, compared with 87.2 and 91.3 when all three primitives are used. The corresponding reductions are 0.1 and 2.1 percentage points, respectively. This experiment supports transfer from the hole training primitive to the evaluated real defect types, without requiring an exact match between the training primitive and test defect type.

\begin{table}[t]
    \centering
    \caption{
        Cross-type supervision results on Real3D-AD.
    }
    \label{tab:pseudo_hole_only_real3d}
    \small
    \begin{tabular}{lcc}
        \toprule
        Training pseudo-anomalies
            & O-AUROC $\uparrow$ & P-AUROC $\uparrow$ \\
        \midrule
        Hole & 87.1 & 89.2 \\
        Bulge, Sink, Hole & 87.2 & 91.3 \\
        \bottomrule
    \end{tabular}
\end{table}

\noindent\textbf{Contribution of Deviation Modeling in CDM.}
CDM uses the clustering procedure described in Section~\ref{implement} to form
feature-space peer groups. Its contribution lies in how these groups
are used for anomaly discrimination, rather than in a new clustering
algorithm. Specifically, $\mathcal{L}_{\mathrm{dev}}$ encourages
anomalous node features to deviate from their assigned cluster
centers, while the deviation embedding
$\mathbf{W}_d\phi(d_g)$ supplies the classifier with cluster-relative
information alongside the EGR-refined node feature.
Thus, the final anomaly score is predicted from the combined
representation, not directly from the cluster deviation.

To examine these two roles, we keep the clustering procedure unchanged and remove the deviation loss and deviation embedding either individually or jointly. Table~\ref{tab:cdm_component_ablation} reports results for in-domain evaluation on Anomaly-ShapeNet and cross-domain evaluation from Anomaly-ShapeNet to Real3D-AD. Removing either component reduces performance in both settings, with a larger cross-domain decrease when the deviation embedding is removed. Removing both components yields the lowest performance among the evaluated configurations. These results support the complementary contributions of supervised deviation shaping and deviation-feature injection within CDM.

\begin{table}[t]
    \centering
    \caption{
        Ablation of the deviation loss and deviation embedding in CDM.
    }
    \label{tab:cdm_component_ablation}
    \small
    \setlength{\tabcolsep}{5pt}
    \begin{tabular}{lcccc}
        \toprule
        \multirow{2.5}{*}{Configuration}
        & \multicolumn{2}{c}{In-Domain ($A \rightarrow A$)}
        & \multicolumn{2}{c}{Cross-Domain ($A \rightarrow R$)}
        \\
        \cmidrule(lr){2-3}
        \cmidrule(lr){4-5}
        & O-AUROC $\uparrow$
        & P-AUROC $\uparrow$
        & O-AUROC $\uparrow$
        & P-AUROC $\uparrow$
        \\
        \midrule
        Full CDM
        & \textbf{97.4}
        & \textbf{94.5}
        & \textbf{83.6}
        & \textbf{89.6}
        \\
        w/o $\mathcal{L}_{\mathrm{dev}}$
        & 96.6 & 92.1 & 79.3 & 84.1
        \\
        w/o deviation emb.
        & 94.9 & 88.7 & 68.5 & 66.8
        \\
        w/o both
        & 94.3 & 86.4 & 59.4 & 53.0
        \\
        \bottomrule
    \end{tabular}
\end{table}

\noindent\textbf{Training Category-Specific Models in Unified Setting.} In the category-specific setting, a separate model is trained and evaluated for each category. In contrast, unified training uses a single shared model across all categories without category labels or category-specific parameters. Although a unified model observes more training samples, it must accommodate heterogeneous normal distributions within a shared representation and anomaly criterion.

To examine the effect of this difference in training settings, we retrain PO3AD~\cite{ye2025po3ad} and PASDF~\cite{zheng2025bridging} under the unified protocol by merging the normal training samples from all categories. Table~\ref{tab:unified_training_comparison} compares the resulting performance with their category-specific counterparts on Real3D-AD and Anomaly-ShapeNet. Both methods exhibit lower O-AUROC and P-AUROC under unified training on both datasets, despite access to more training samples. These results indicate that pooling training data across categories does not automatically improve performance for the evaluated methods, highlighting the challenge of representing heterogeneous normal structures within a single shared model.

\begin{table}[t]
    \centering
    \caption{
        Comparison between category-specific and unified training on Real3D-AD and Anomaly-ShapeNet.(C) denotes the category-specific setting and (U) denotes the unified setting.
    }
    \label{tab:unified_training_comparison}
    \small
    \setlength{\tabcolsep}{4pt}
    \begin{tabular}{lcccc}
        \toprule
        &
        \multicolumn{2}{c}{Real3D-AD}
        &
        \multicolumn{2}{c}{Anomaly-ShapeNet}
        \\
        \cmidrule(lr){2-3}
        \cmidrule(lr){4-5}
        Method
        & O-AUROC $\uparrow$
        & P-AUROC $\uparrow$
        & O-AUROC $\uparrow$
        & P-AUROC $\uparrow$
        \\
        \midrule
        PO3AD (C)
        & \textbf{76.5} & \textbf{65.0} & \textbf{83.9} & \textbf{89.8}
        \\
        PO3AD (U)
        & 54.7 & 62.2 & 72.3 & 75.1
        \\
        \midrule
        PASDF (C)
        & \textbf{80.2} & \textbf{74.5} & \textbf{90.0} & \textbf{89.7}
        \\
        PASDF (U)
        & 64.7 & 58.4 & 77.5 & 73.6
        \\
        \bottomrule
    \end{tabular}
\end{table}

\noindent\textbf{Additional Cross-Domain Comparison under Unified Training.} We further evaluate PO3AD~\cite{ye2025po3ad} and PASDF~\cite{zheng2025bridging} under the unified cross-domain setting from Anomaly-ShapeNet to Real3D-AD. For each method, normal training samples from all Anomaly-ShapeNet categories are merged to train a single shared model without category labels. The source-trained model is then applied directly to Real3D-AD without fine-tuning or domain-specific adaptation. 

Table~\ref{tab:unified_crossdomain_comparison} reports the results. Our method outperforms both additional unified baselines in object-level detection and point-level localization. This comparison further strengthens our claim of superior cross-domain performance.

\begin{table}[t]
    \centering
    \caption{
        Additional unified cross-domain comparison. (U) denotes the unified setting.
    }
    \label{tab:unified_crossdomain_comparison}
    \small
    \setlength{\tabcolsep}{8pt}
    \begin{tabular}{lcc}
        \toprule
        Method
        & O-AUROC $\uparrow$
        & P-AUROC $\uparrow$
        \\
        \midrule
        PO3AD (U)
        & 52.2 & 48.7
        \\
        PASDF (U)
        & 50.8 & 43.4
        \\
        \midrule
        \rowcolor{pubcolor!25}
        Ours
        & \textbf{83.6} & \textbf{89.6}
        \\
        \bottomrule
    \end{tabular}

    \smallskip
    \begin{minipage}{0.95\linewidth}
    \end{minipage}
\end{table}


\subsection{Limitations and Future Works}
\label{limitation}
\noindent\textbf{Overall.} Our framework operates solely on point cloud data, which is an intentional design choice to focus on geometric relation-based defect detection. However, in practical industrial environments, RGB-D cameras naturally capture color and depth information alongside point clouds. Such complementary modalities can provide additional defect cues — such as surface color variations or subtle depth discontinuities — that are difficult to capture from geometry alone. Integrating multi-modal information into our relational inconsistency modeling framework therefore represents a promising direction for further improving detection performance, which we leave as future work.

\noindent\textbf{Anomaly Sparsity in CDM.} The cluster-relative cue in CDM relies on anomalies being sparse relative to normal structures. If an anomalous region is sufficiently large and coherent, its features may form a separate cluster, making their deviations from the resulting cluster center small. Consequently, the CDM signal may weaken under this condition. Although the classifier also receives EGR-refined node features, this additional input does not by itself establish robustness to large coherent anomalies. Reducing the number of clusters to form larger peer groups may be a possible direction for addressing this problem.


\subsection{Additional Quantitative Results}
\label{Quantitative}
Tables~\ref{tab:table8}--\ref{tab:table9} and~\ref{tab:table10}--\ref{tab:table11} report per-category O-AUROC and P-AUROC on Anomaly-ShapeNet and Real3D-AD, respectively. For cross-domain evaluation, Tables~\ref{tab:table12}--\ref{tab:table13} cover the case of training on Anomaly-ShapeNet and testing on Real3D-AD, and Tables~\ref{tab:table14}--\ref{tab:table15} cover the reverse direction. Table~\ref{tab:table16} additionally reports per-category object-level AUPR on Anomaly-ShapeNet.

\subsection{Additional Qualitative Results}
\label{Qualitative}
\begin{figure*}[!t] 
\centerline{\includegraphics[width=0.9\linewidth]{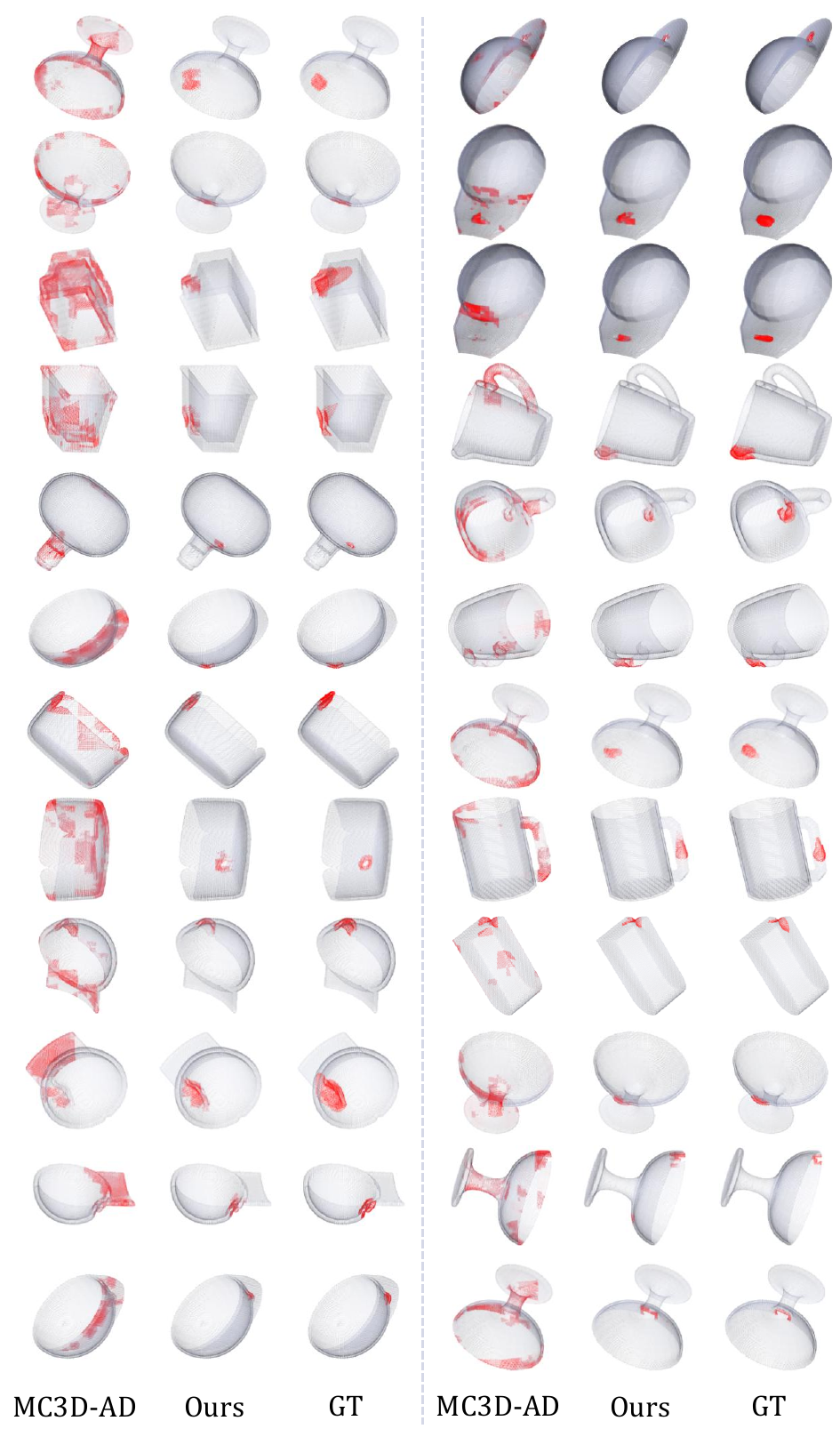}}
    \caption{Additional in-domain qualitative comparison with MC3D-AD on Anomaly-ShapeNet.}
    \label{fig:supp_as_1} 
\end{figure*}

\begin{figure*}[!t] 
\centerline{\includegraphics[width=0.9\linewidth]{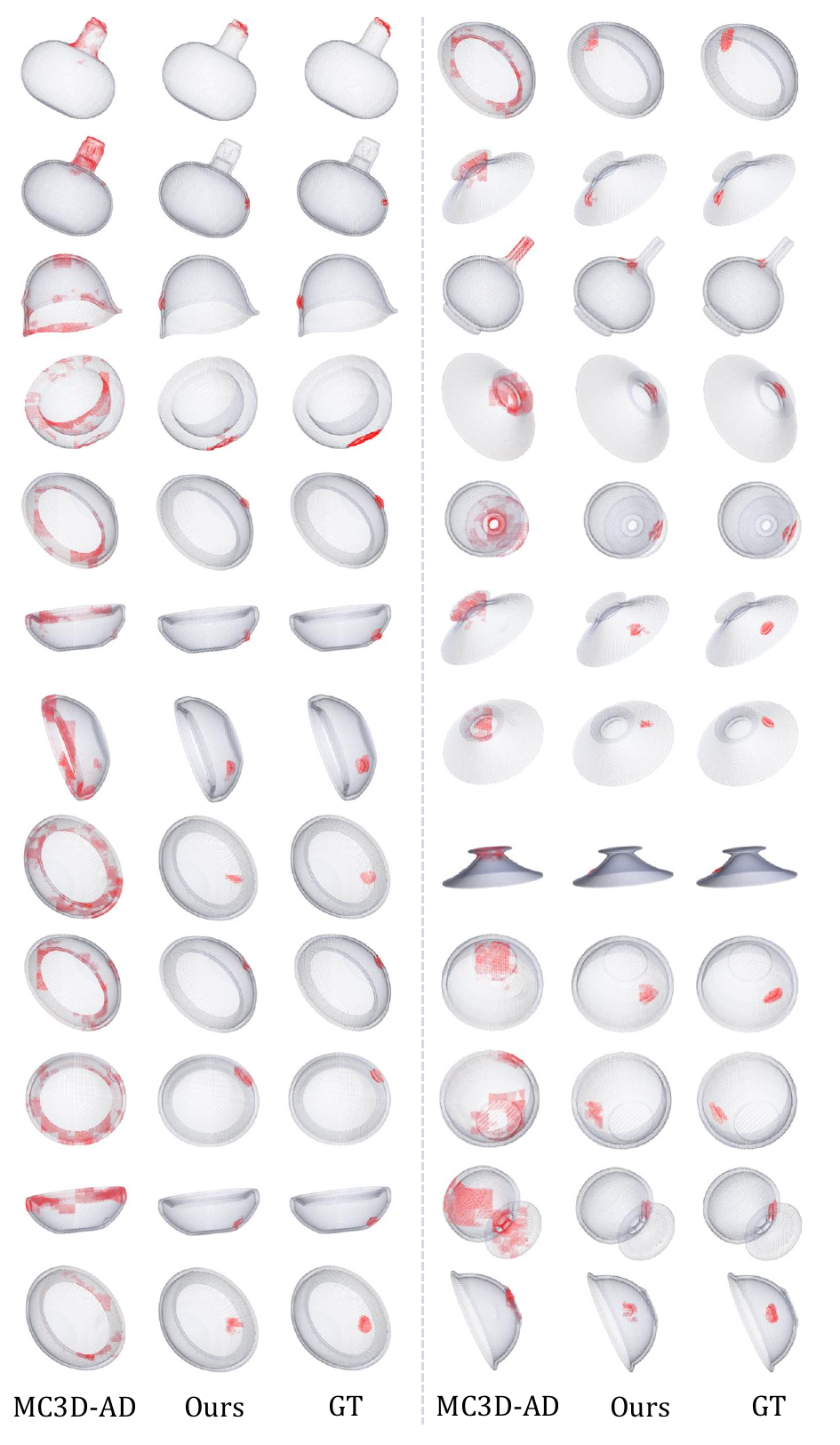}}
    \caption{Additional in-domain qualitative comparison with MC3D-AD on Anomaly-ShapeNet.}
    \label{fig:supp_as_2} 
\end{figure*}

\begin{figure*}[!t] 
\centerline{\includegraphics[width=0.9\linewidth]{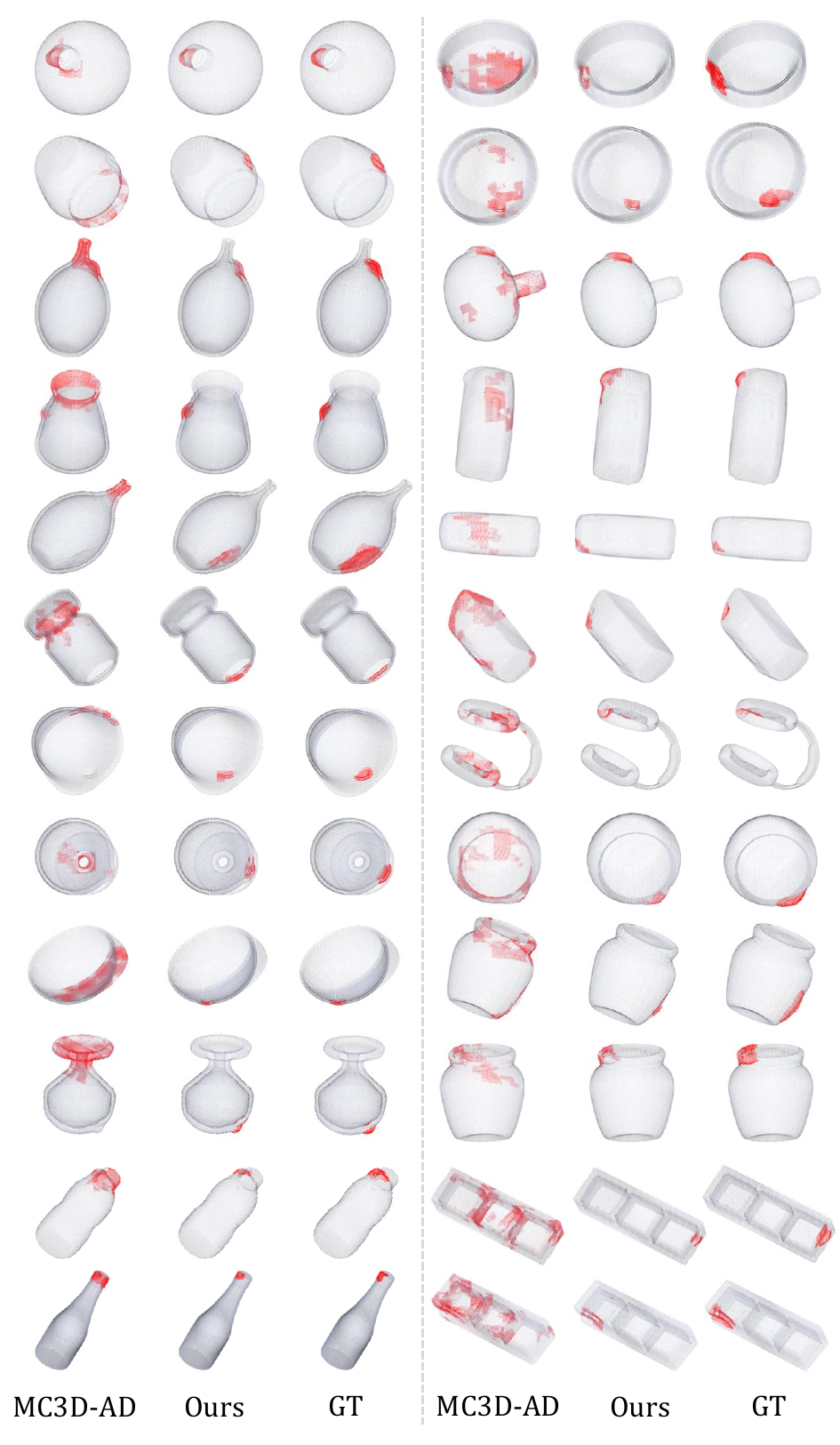}}
    \caption{Additional in-domain qualitative comparison with MC3D-AD on Anomaly-ShapeNet.}
    \label{fig:supp_as_3} 
\end{figure*}

\begin{figure*}[h] 
\centerline{\includegraphics[width=0.9\linewidth]{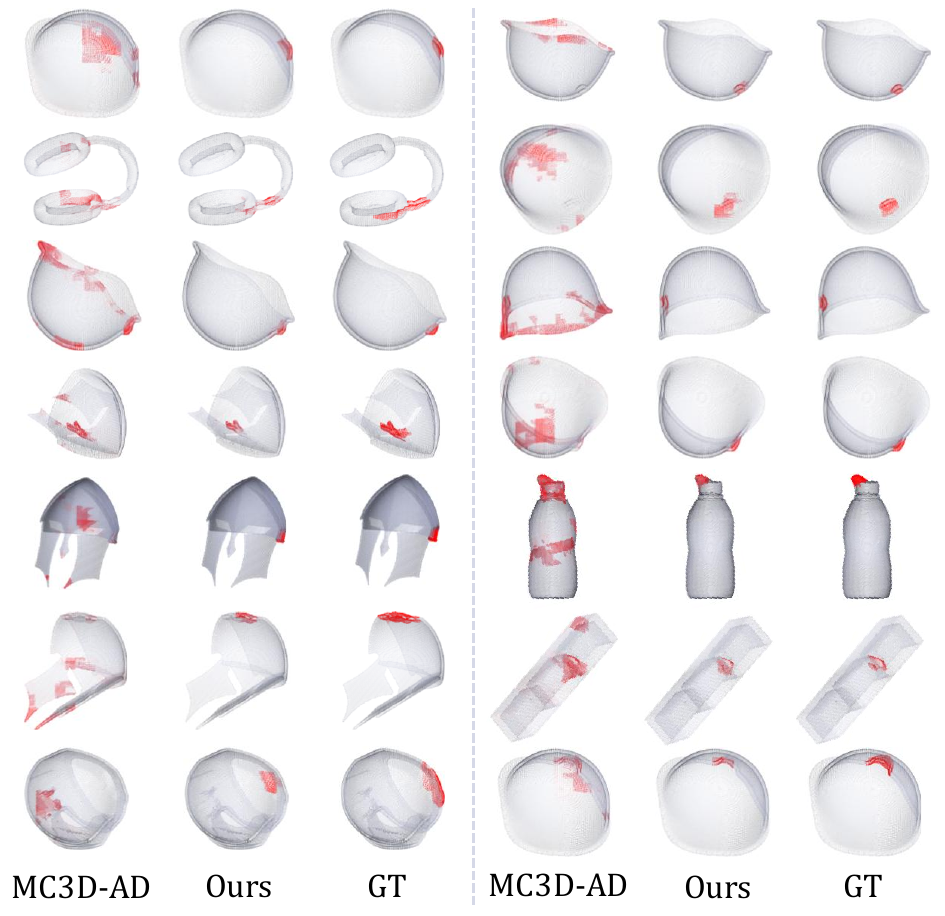}}
    \caption{Additional in-domain qualitative comparison with MC3D-AD on Anomaly-ShapeNet.}
    \label{fig:supp_as_4} 
\end{figure*}

\begin{figure*}[!t] 
\centerline{\includegraphics[width=0.9\linewidth]{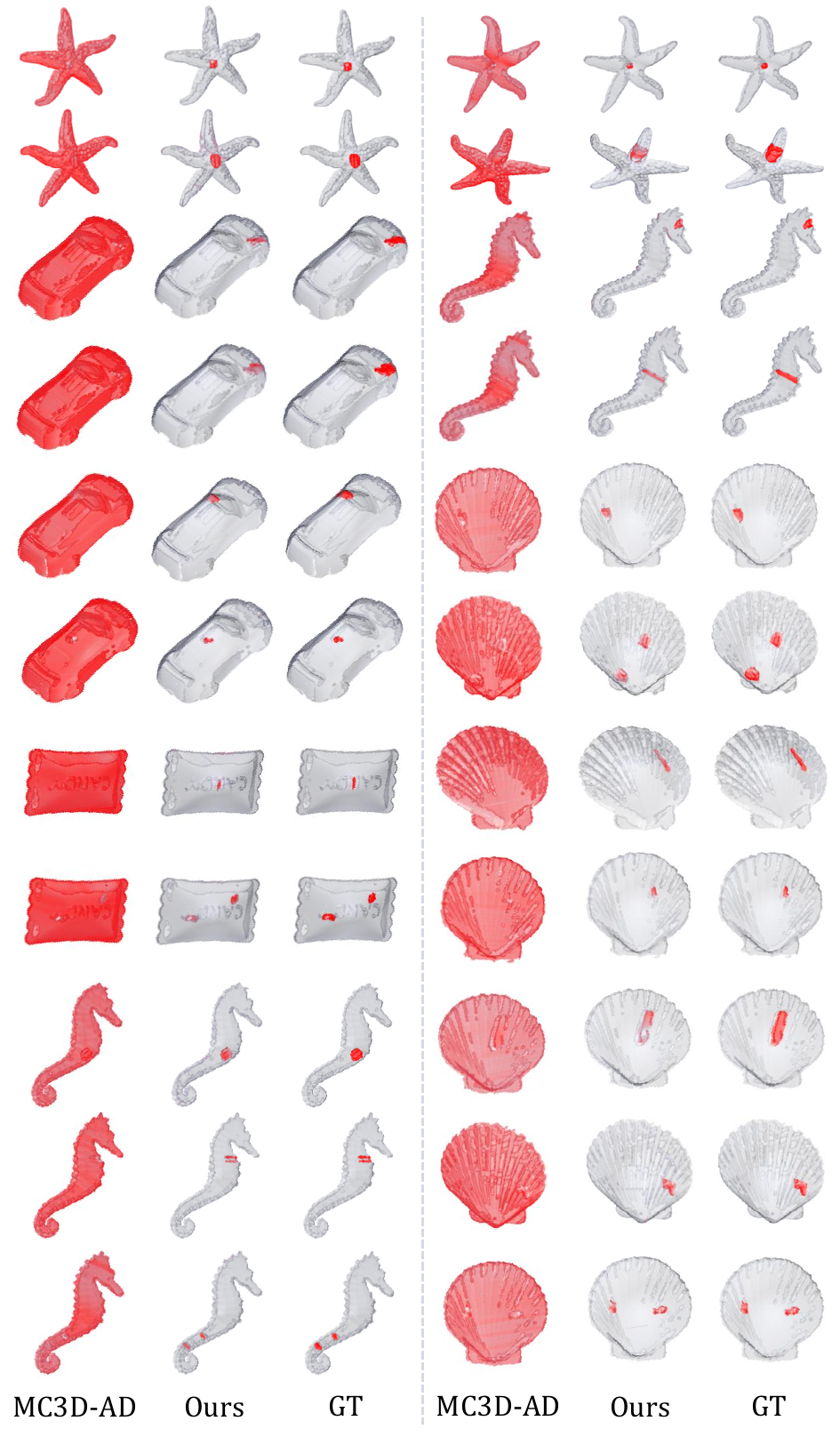}}
    \caption{Additional cross-domain qualitative comparison with MC3D-AD. Models are trained on Anomaly-ShapeNet and tested on Real3D-AD.}
    \label{fig:supp_real3d_1} 
\end{figure*}

\begin{figure*}[!t] 
\centerline{\includegraphics[width=0.9\linewidth]{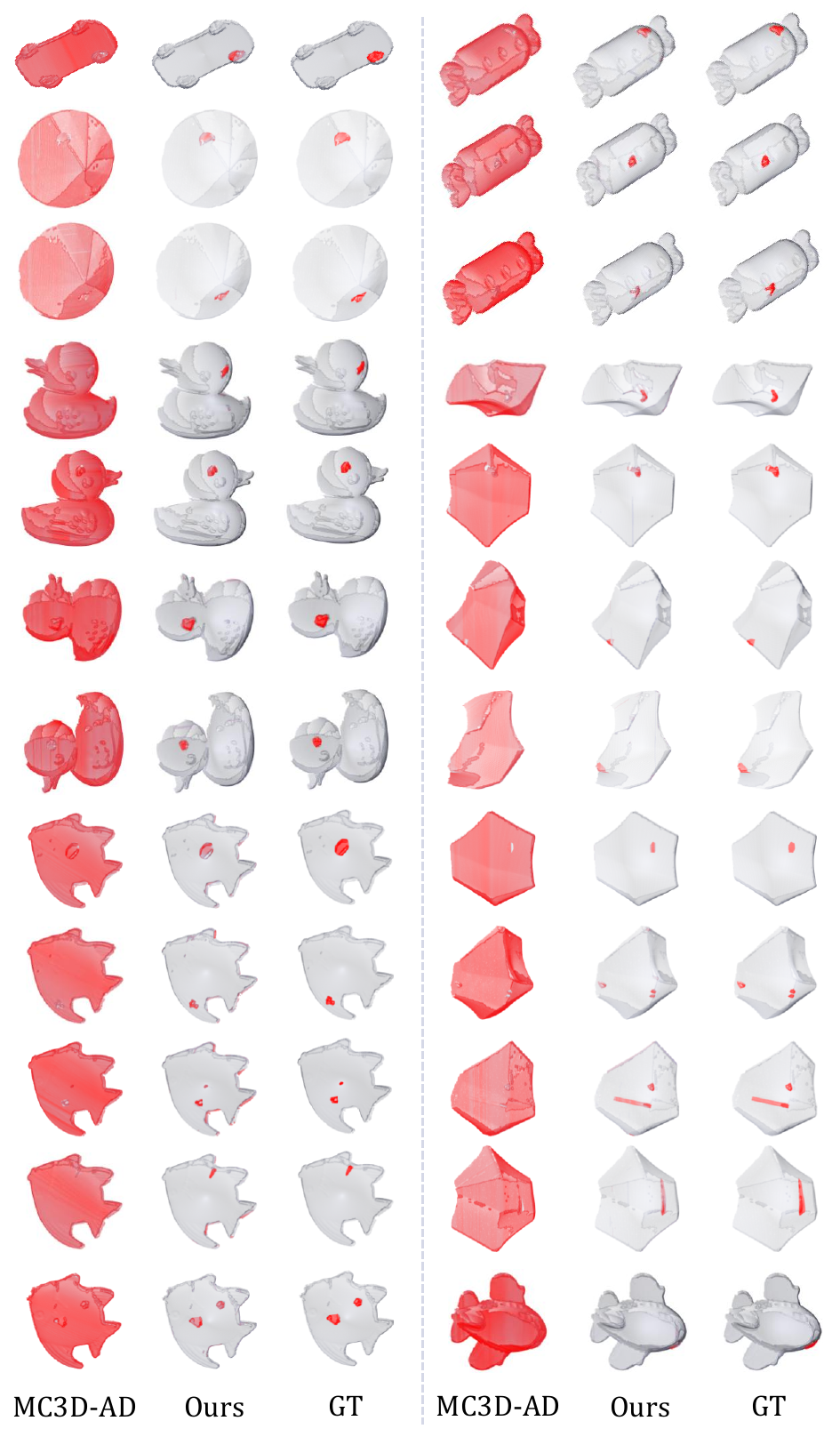}}
    \caption{Additional cross-domain qualitative comparison with MC3D-AD. Models are trained on Anomaly-ShapeNet and tested on Real3D-AD.}
    \label{fig:supp_real3d_2} 
\end{figure*}

Figures~\ref{fig:supp_as_1}--\ref{fig:supp_as_4} present additional in-domain qualitative results across diverse categories of Anomaly-ShapeNet, and Figures~\ref{fig:supp_real3d_1}--\ref{fig:supp_real3d_2} present cross-domain results trained on Anomaly-ShapeNet and tested on Real3D-AD.


\begin{table*}[!h]
\caption{Comparison of object-level AUROC (\%) across all categories on Anomaly-ShapeNet.}
\resizebox{\textwidth}{!}{
\begin{tabular}{c|l|cccccccccccccc}
\toprule
\multicolumn{16}{c}{O-AUROC($\uparrow$)}\\
\midrule
\multicolumn{2}{c|}{Method} & ashtray0 & bag0 & bottle0 & bottle1 & bottle3 & bowl0 & bowl1 & bowl2 & bowl3 & bowl4 & bowl5 & bucket0 & bucket1 & cap0 \\
\midrule
\multirow{12}{*}{\shortstack{Category-\\Specific}}
& BTF(Raw)            & 57.8 & 41.0 & 59.7 & 51.0 & 56.8 & 56.4 & 26.4 & 52.5 & 38.5 & 66.4 & 41.7 & 61.7 & 32.1 & 66.8 \\
& BTF(FPFH)           & 42.0 & 54.6 & 34.4 & 54.6 & 32.2 & 50.9 & 66.8 & 51.0 & 49.0 & 60.9 & 69.9 & 40.1 & 63.3 & 61.8 \\
& M3DM                & 57.7 & 53.7 & 57.4 & 63.7 & 54.1 & 63.4 & 66.3 & 68.4 & 61.7 & 46.4 & 40.9 & 30.9 & 50.1 & 55.7 \\
& PatchCore(FPFH)     & 58.7 & 57.1 & 60.4 & 66.7 & 57.2 & 50.4 & 63.9 & 61.5 & 53.7 & 49.4 & 55.8 & 46.9 & 55.1 & 58.0 \\
& PatchCore(PointMAE) & 59.1 & 60.1 & 51.3 & 60.1 & 65.0 & 52.3 & 62.9 & 45.8 & 57.9 & 50.1 & 59.3 & 59.3 & 56.1 & 58.9 \\
& CPMF                & 35.3 & 64.3 & 52.0 & 48.2 & 40.5 & 78.3 & 63.9 & 62.5 & 65.8 & 68.3 & 68.5 & 48.2 & 60.1 & 60.1 \\
& Reg3D-AD            & 59.7 & 70.6 & 48.6 & 69.5 & 52.5 & 67.1 & 52.5 & 49.0 & 34.8 & 66.3 & 59.3 & 61.0 & 75.2 & 69.3 \\
& IMRNet              & 67.1 & 66.0 & 55.2 & 70.0 & 64.0 & 68.1 & 70.2 & 68.5 & 59.9 & 67.6 & {71.0} & 58.0 & {77.1} & 73.7 \\
& R3D-AD              & {83.3} & {72.0} & {73.3} & {73.7} & {78.1} & {81.9} & {77.8} & {74.1} & {76.7} & {74.4} & 65.6 & {68.3} & 75.6 & {82.2} \\
& PO3AD               & \textbf{100.0} & {83.3} & {90.0} & {93.3} & {92.6} & {92.2} & {82.9} & {83.3} & {88.1} & \textbf{98.1} & {84.9} & {85.3} & {78.7} & \underline{87.7} \\
& PASDF               & \textbf{100.0} & \underline{99.5} & \textbf{100.0} & \textbf{100.0} & \textbf{100.0} & \textbf{100.0} & {94.8} & \textbf{100.0} & \textbf{100.0} & {93.3} & \underline{91.2} & \textbf{96.8} & {77.5} & {85.2} \\
& Reg2Inv             & {90.0} & \textbf{100.0} & \textbf{100.0} & \textbf{100.0} & \textbf{100.0} & \textbf{100.0} & {80.7} & {65.6} & {58.5} & {85.2} & {81.8} & {81.3} & \underline{90.2} & {65.9} \\
\midrule[1.2pt]
\multirow{2}{*}{\textbf{Unified}}
& MC3D-AD             & {96.2} & {80.5} & {79.5} & {70.9} & {75.6} & {93.0} & \underline{97.8} & {71.9} & {88.5} & {91.1} & {75.4} & {89.8} & {78.4} & {79.3} \\
& \cellcolor{pubcolor!25}\textbf{GRIM (Ours)} & \cellcolor{pubcolor!25}\textbf{100.0} & \cellcolor{pubcolor!25}{97.1} & \cellcolor{pubcolor!25}\textbf{100.0} & \cellcolor{pubcolor!25}{99.3} & \cellcolor{pubcolor!25}{94.3} & \cellcolor{pubcolor!25}{95.9} & \cellcolor{pubcolor!25}\textbf{100.0} & \cellcolor{pubcolor!25}\underline{93.3} & \cellcolor{pubcolor!25}\underline{97.8} & \cellcolor{pubcolor!25}\underline{96.7} & \cellcolor{pubcolor!25}\textbf{98.9} & \cellcolor{pubcolor!25}\underline{95.9} & \cellcolor{pubcolor!25}\textbf{99.4} & \cellcolor{pubcolor!25}\textbf{98.9} \\
\bottomrule
\end{tabular}
}
\resizebox{\textwidth}{!}{
\begin{tabular}{c|l|ccccccccccccc}
\toprule
\multicolumn{2}{c|}{Method} & cap3 & cap4 & cap5 & cup0 & cup1 & eraser0 & headset0 & headset1 & helmet0 & helmet1 & helmet2 & helmet3 & jar \\
\midrule
\multirow{12}{*}{\shortstack{Category-\\Specific}}
& BTF(Raw)            & 52.7 & 46.8 & 37.3 & 40.3 & 52.1 & 52.5 & 37.8 & 51.5 & 55.3 & 34.9 & 60.2 & 52.6 & 42.0 \\
& BTF(FPFH)           & 52.2 & 52.0 & 58.6 & 58.6 & 61.0 & 71.9 & 52.0 & 49.0 & 57.1 & 71.9 & 54.2 & 44.4 & 58.6 \\
& M3DM                & 42.3 & 77.7 & 63.9 & 53.9 & 55.6 & 62.7 & 57.7 & 61.7 & 52.6 & 42.7 & 62.3 & 37.4 & 56.4 \\
& PatchCore(FPFH)     & 45.3 & 75.7 & 79.0 & 60.0 & 58.6 & 65.7 & 58.3 & 63.7 & 54.6 & 48.4 & 42.5 & 40.4 & 49.4 \\
& PatchCore(PointMAE) & 47.6 & 72.7 & 53.8 & 61.0 & 55.6 & 67.7 & 59.1 & 62.7 & 55.6 & 55.2 & 44.7 & 42.4 & 48.3 \\
& CPMF                & 55.1 & 55.3 & {69.7} & 49.7 & 49.9 & 68.9 & 64.3 & 45.8 & 55.5 & 58.9 & 46.2 & 52.0 & 61.0 \\
& Reg3D-AD            & 72.5 & 64.3 & 46.7 & 51.0 & 53.8 & 34.3 & 53.7 & 61.0 & 60.0 & 38.1 & 61.4 & 36.7 & 59.2 \\
& IMRNet              & {77.5} & 65.2 & 65.2 & 64.3 & {75.7} & 54.8 & 72.0 & 67.6 & 59.7 & 60.0 & {64.1} & 57.3 & 78.0 \\
& R3D-AD              & 73.0 & {68.1} & {67.0} & {77.6} & {75.7} & {89.0} & {73.8} & {79.5} & {75.7} & {72.0} & 63.3 & {70.7} & {83.8} \\
& PO3AD               & {85.9} & {79.2} & {67.0} & {87.1} & {83.3} & \underline{99.5} & {80.8} & \underline{92.3} & {76.2} & {96.1} & {86.9} & {75.4} & {86.6} \\
& PASDF               & {64.9} & {64.6} & {85.3} & \underline{97.1} & {85.7} & {95.2} & \textbf{100.0} & {79.5} & {81.2} & {93.8} & {76.5} & {84.6} & \textbf{100.0} \\
& Reg2Inv             & \underline{86.3} & {68.1} & \underline{90.2} & {73.3} & {93.3} & \textbf{100.0} & \textbf{100.0} & {84.3} & \underline{81.7} & {98.6} & \underline{87.5} & {87.6} & \textbf{100.0} \\
\midrule[1.2pt]
\multirow{2}{*}{\textbf{Unified}}
& MC3D-AD             & {70.1} & \underline{83.5} & {76.1} & {74.3} & \underline{95.2} & {77.6} & {86.2} & {88.6} & {67.2} & \textbf{100.0} & {60.9} & \textbf{97.9} & {97.1} \\
& \cellcolor{pubcolor!25}\textbf{GRIM (Ours)} & \cellcolor{pubcolor!25}\textbf{99.6} & \cellcolor{pubcolor!25}\textbf{98.6} & \cellcolor{pubcolor!25}\textbf{98.2} & \cellcolor{pubcolor!25}\textbf{98.6} & \cellcolor{pubcolor!25}\textbf{100.0} & \cellcolor{pubcolor!25}{84.8} & \cellcolor{pubcolor!25}{98.2} & \cellcolor{pubcolor!25}\textbf{96.2} & \cellcolor{pubcolor!25}\textbf{97.7} & \cellcolor{pubcolor!25}\underline{99.0} & \cellcolor{pubcolor!25}\textbf{99.4} & \cellcolor{pubcolor!25}\underline{94.8} & \cellcolor{pubcolor!25}\textbf{100.0} \\
\bottomrule
\end{tabular}
}
\resizebox{\textwidth}{!}{
\begin{tabular}{c|l|ccccccccccccc|c}
\toprule
\multicolumn{2}{c|}{Method} & phone & shelf0 & tap0 & tap1 & vase0 & vase1 & vase2 & vase3 & vase4 & vase5 & vase7 & vase8 & vase9 & Mean \\
\midrule
\multirow{12}{*}{\shortstack{Category-\\Specific}}
& BTF(Raw)            & 56.3 & 16.4 & 52.5 & 57.3 & 53.1 & 54.9 & 41.0 & 71.7 & 42.5 & 58.5 & 44.8 & 42.4 & 56.4 & 49.3 \\
& BTF(FPFH)           & 67.1 & 60.9 & 56.0 & 54.6 & 34.2 & 21.9 & 54.6 & 69.9 & 51.0 & 40.9 & 51.8 & 66.8 & 26.8 & 52.8 \\
& M3DM                & 35.7 & 56.4 & {75.4} & 73.9 & 57.4 & 42.7 & 73.7 & 43.9 & 47.6 & 43.9 & 65.7 & 66.3 & 66.3 & 55.2 \\
& PatchCore(FPFH)     & 38.8 & 49.4 & {75.3} & {76.6} & 60.4 & 42.3 & 72.1 & 44.9 & 50.6 & 44.9 & 69.3 & 66.2 & 66.0 & 56.8 \\
& PatchCore(PointMAE) & 48.8 & 52.3 & 45.8 & 53.8 & 51.3 & 55.2 & 74.1 & 46.0 & 51.6 & 46.0 & 65.0 & 66.3 & 62.9 & 56.2 \\
& CPMF                & 50.9 & 68.5 & 35.9 & 69.7 & 58.2 & 34.5 & 58.2 & 58.2 & 51.4 & 58.2 & 39.7 & 52.9 & 60.9 & 55.9 \\
& Reg3D-AD            & 41.4 & {68.8} & 67.6 & 64.1 & 48.6 & 70.2 & 60.5 & 65.0 & 50.0 & 65.0 & 46.2 & 62.0 & 59.4 & 57.2 \\
& IMRNet              & 75.5 & 60.3 & 67.6 & 69.6 & 55.2 & {75.7} & 61.4 & 70.0 & 52.4 & 70.0 & 63.5 & 63.0 & 59.4 & 66.1 \\
& R3D-AD              & {76.2} & {69.6} & 73.6 & {90.0} & {73.3} & 72.9 & {75.2} & {74.2} & {63.0} & {74.2} & {77.1} & {72.1} & {71.8} & {74.9} \\
& PO3AD               & {77.6} & 57.3 & 74.5 & 68.1 & {85.8} & {74.2} & {95.2} & {82.1} & {67.5} & {85.2} & {96.6} & {73.9} & {83.0} & {83.9} \\
& PASDF               & \textbf{100.0} & 71.3 & 88.2 & 79.3 & \textbf{100.0} & \underline{92.9} & \textbf{100.0} & {80.6} & \underline{91.2} & \textbf{100.0} & \textbf{100.0} & \underline{92.4} & {83.6} & \underline{90.0} \\
& Reg2Inv             & \textbf{100.0} & 57.7 & \underline{94.8} & 80.4 & {99.6} & {60.5} & \textbf{100.0} & \underline{84.5} & {81.8} & \textbf{100.0} & {64.3} & {81.8} & \underline{87.3} & {86.1} \\
\midrule[1.2pt]
\multirow{2}{*}{\textbf{Unified}}
& MC3D-AD             & {91.9} & \underline{84.1} & {94.5} & \textbf{97.0} & {79.5} & {85.7} & {92.9} & {76.1} & {87.6} & {76.1} & {93.8} & {67.0} & {73.6} & {84.2} \\
& \cellcolor{pubcolor!25}\textbf{GRIM (Ours)} & \cellcolor{pubcolor!25}{95.2} & \cellcolor{pubcolor!25}\textbf{95.4} & \cellcolor{pubcolor!25}\textbf{99.4} & \cellcolor{pubcolor!25}\underline{91.1} & \cellcolor{pubcolor!25}\textbf{100.0} & \cellcolor{pubcolor!25}\textbf{97.6} & \cellcolor{pubcolor!25}{95.2} & \cellcolor{pubcolor!25}\textbf{97.6} & \cellcolor{pubcolor!25}\textbf{99.4} & \cellcolor{pubcolor!25}\textbf{100.0} & \cellcolor{pubcolor!25}\textbf{100.0} & \cellcolor{pubcolor!25}\textbf{95.8} & \cellcolor{pubcolor!25}\textbf{96.7} & \cellcolor{pubcolor!25}\textbf{97.4} \\
\bottomrule
\end{tabular}
}
\label{tab:table8}
\vspace{-4mm}
\end{table*}

\begin{table*}[t!]
\caption{Comparison of point-level AUROC (\%) across all categories on Anomaly-ShapeNet.}
\resizebox{\textwidth}{!}{
\begin{tabular}{c|l|ccccccccccccc}
\toprule
\multicolumn{15}{c}{P-AUROC($\uparrow$)}\\
\midrule
\multicolumn{2}{c|}{Method} & ashtray0 & bag0 & bottle0 & bottle1 & bottle3 & bowl0 & bowl1 & bowl2 & bowl3 & bowl4 & bowl5 & bucket0 & bucket1 \\
\midrule
\multirow{11}{*}{\shortstack{Category-\\Specific}}
& BTF(Raw)            & 51.2 & 43.0 & 55.1 & 49.1 & {72.0} & 52.4 & 46.4 & 42.6 & {68.5} & 56.3 & 51.7 & 61.7 & 68.6 \\
& BTF(FPFH)           & 62.4 & {74.6} & 64.1 & 54.9 & 62.2 & 71.0 & {76.8} & 51.8 & 59.0 & 67.9 & 69.9 & 40.1 & 63.3 \\
& M3DM                & 57.7 & 63.7 & 66.3 & 63.7 & 53.2 & 65.8 & 66.3 & {69.4} & 65.7 & 62.4 & 48.9 & {69.8} & 69.9 \\
& PatchCore(FPFH)     & 59.7 & 57.4 & 65.4 & 68.7 & 51.2 & 52.4 & 53.1 & 62.5 & 32.7 & 72.0 & 35.8 & 45.9 & 57.1 \\
& PatchCore(PointMAE) & 49.5 & 67.4 & 55.3 & 60.6 & 65.3 & 52.7 & 52.4 & 51.5 & 58.1 & 50.1 & 56.2 & 58.6 & 57.4 \\
& CPMF                & 61.5 & 65.5 & 52.1 & 57.1 & 43.5 & 74.5 & 48.8 & 63.5 & 64.1 & 68.3 & 68.4 & 48.6 & 60.1 \\
& Reg3D-AD            & {69.8} & 71.5 & {88.6} & 69.6 & 52.5 & 77.5 & 61.5 & 59.3 & 65.4 & {80.0} & 69.1 & 61.9 & 75.2 \\
& IMRNet              & 67.1 & 66.8 & 55.6 & {70.2} & 64.1 & {78.1} & 70.5 & 68.4 & 59.9 & 57.6 & {71.5} & 58.5 & {77.4} \\
& PO3AD               & \textbf{96.2} & {94.9} & {91.2} & {84.4} & {88.0} & \underline{97.8} & \underline{91.4} & \textbf{91.8} & {93.5} & \underline{96.7} & \underline{94.1} & {75.5} & \underline{89.9} \\
& PASDF               & \underline{91.9} & \underline{95.8} & {95.1} & \underline{92.6} & \underline{94.8} & {96.3} & {90.0} & {81.6} & \underline{93.9} & {86.5} & {90.9} & \underline{87.5} & {82.4} \\
& Reg2Inv             & {78.5} & \textbf{99.1} & \textbf{99.5} & {84.9} & {81.7} & \textbf{98.3} & {82.8} & {82.2} & {76.1} & {78.8} & {82.4} & {61.0} & {85.5} \\
\midrule[1.2pt]
\multirow{2}{*}{\textbf{Unified}}
& MC3D-AD             & {80.1} & {81.5} & {89.5} & {88.3} & {90.1} & {82.7} & {53.4} & {60.7} & {78.3} & {65.4} & {55.2} & {79.0} & {89.5} \\
& \cellcolor{pubcolor!25}\textbf{GRIM (Ours)} & \cellcolor{pubcolor!25}{89.9} & \cellcolor{pubcolor!25}{95.0} & \cellcolor{pubcolor!25}\underline{98.1} & \cellcolor{pubcolor!25}\textbf{94.7} & \cellcolor{pubcolor!25}\textbf{97.4} & \cellcolor{pubcolor!25}{96.5} & \cellcolor{pubcolor!25}\textbf{92.5} & \cellcolor{pubcolor!25}\underline{90.0} & \cellcolor{pubcolor!25}\textbf{95.6} & \cellcolor{pubcolor!25}\textbf{97.1} & \cellcolor{pubcolor!25}\textbf{96.5} & \cellcolor{pubcolor!25}\textbf{89.9} & \cellcolor{pubcolor!25}\textbf{95.1} \\
\bottomrule
\end{tabular}
}
\resizebox{\textwidth}{!}{
\begin{tabular}{c|l|ccccccccccccc}
\toprule
\multicolumn{2}{c|}{Method} & cap0 & cap3 & cap4 & cap5 & cup0 & cup1 & eraser0 & headset0 & headset1 & helmet0 & helmet1 & helmet2 & helmet3 \\
\midrule
\multirow{11}{*}{\shortstack{Category-\\Specific}}
& BTF(Raw)            & 52.4 & 68.7 & 46.9 & 37.3 & 63.2 & 56.1 & 63.7 & 57.8 & 47.5 & 50.4 & 44.9 & 60.5 & 70.0 \\
& BTF(FPFH)           & {73.0} & 65.8 & 52.4 & 58.6 & {79.0} & 61.9 & 71.9 & 62.0 & 59.1 & 57.5 & {74.9} & 64.3 & 72.4 \\
& M3DM                & 53.1 & 60.5 & 71.8 & 65.5 & 71.5 & 55.6 & 71.0 & 58.1 & 58.5 & 59.9 & 42.7 & 62.3 & 65.5 \\
& PatchCore(FPFH)     & 47.2 & 65.3 & 59.5 & {79.5} & 65.5 & 59.6 & {81.0} & 58.3 & 46.4 & 54.8 & 48.9 & 45.5 & {73.7} \\
& PatchCore(PointMAE) & 54.4 & 48.8 & 72.5 & 54.5 & 51.0 & {85.6} & 37.8 & 57.5 & 42.3 & 58.0 & 56.2 & 65.1 & 61.5 \\
& CPMF                & 60.1 & 55.1 & 55.3 & 55.1 & 49.7 & 50.9 & 68.9 & 69.9 & 45.8 & 55.5 & 54.2 & 51.5 & 52.0 \\
& Reg3D-AD            & 63.2 & {71.8} & {81.5} & 46.7 & 68.5 & 69.8 & 75.5 & 58.0 & {62.6} & {60.0} & 62.4 & {82.5} & 62.0 \\
& IMRNet              & 71.5 & 70.6 & 75.3 & 74.2 & 64.3 & 68.8 & 54.8 & {70.5} & 47.6 & 59.8 & 60.4 & 64.4 & 66.3 \\
& PO3AD               & \underline{95.7} & \underline{94.8} & \underline{94.0} & {86.4} & {90.9} & \underline{93.2} & \underline{97.4} & {82.3} & {90.7} & {87.8} & \textbf{94.8} & \underline{93.2} & {84.6} \\
& PASDF               & {94.8} & {86.1} & {89.4} & {92.0} & \underline{94.8} & {88.4} & {94.5} & {86.3} & {89.1} & {81.6} & {64.6} & {80.9} & \textbf{95.8} \\
& Reg2Inv             & {86.1} & {94.5} & {86.4} & \textbf{97.0} & {79.8} & {88.1} & \textbf{98.0} & \textbf{94.6} & \textbf{97.0} & \textbf{92.5} & \underline{90.6} & {89.1} & \underline{95.6} \\
\midrule[1.2pt]
\multirow{2}{*}{\textbf{Unified}}
& MC3D-AD             & {86.7} & {92.2} & {88.0} & {88.5} & {82.9} & {71.5} & {85.6} & {66.6} & {70.9} & {74.4} & {58.2} & {81.7} & {60.2} \\
& \cellcolor{pubcolor!25}\textbf{GRIM (Ours)} & \cellcolor{pubcolor!25}\textbf{97.8} & \cellcolor{pubcolor!25}\textbf{97.6} & \cellcolor{pubcolor!25}\textbf{94.3} & \cellcolor{pubcolor!25}\underline{94.3} & \cellcolor{pubcolor!25}\textbf{97.5} & \cellcolor{pubcolor!25}\textbf{97.0} & \cellcolor{pubcolor!25}{90.4} & \cellcolor{pubcolor!25}\underline{92.9} & \cellcolor{pubcolor!25}\underline{93.8} & \cellcolor{pubcolor!25}\underline{90.7} & \cellcolor{pubcolor!25}{88.1} & \cellcolor{pubcolor!25}\textbf{95.1} & \cellcolor{pubcolor!25}{88.7} \\
\bottomrule
\end{tabular}
}
\resizebox{\textwidth}{!}{
\begin{tabular}{c|l|cccccccccccccc|c}
\toprule
\multicolumn{2}{c|}{Method} & jar & phone & shelf0 & tap0 & tap1 & vase0 & vase1 & vase2 & vase3 & vase4 & vase5 & vase7 & vase8 & vase9 & Mean \\
\midrule
\multirow{11}{*}{\shortstack{Category-\\Specific}}
& BTF(Raw)            & 42.3 & 58.3 & 46.4 & 52.7 & 56.4 & 61.8 & 54.9 & 40.3 & 60.2 & 61.3 & 58.5 & 57.8 & 55.0 & 56.4 & 55.0 \\
& BTF(FPFH)           & 42.7 & 67.5 & 61.9 & 56.8 & 59.6 & 64.2 & 61.9 & 64.6 & {69.9} & 71.0 & 42.9 & 54.0 & 66.2 & 56.8 & 62.8 \\
& M3DM                & 54.1 & 35.8 & 55.4 & 65.4 & 71.2 & 60.8 & 60.2 & 73.7 & 65.8 & 65.5 & 64.2 & 51.7 & 55.1 & 66.3 & 61.6 \\
& PatchCore(FPFH)     & 47.8 & 48.8 & 61.3 & 73.3 & {76.8} & 65.5 & 45.3 & 72.1 & 43.0 & 50.5 & 44.7 & 69.3 & 57.5 & 66.3 & 58.0 \\
& PatchCore(PointMAE) & 48.7 & {88.6} & 54.3 & {85.8} & 54.1 & {67.7} & 55.1 & {74.2} & 46.5 & 52.3 & 57.2 & 65.1 & 36.4 & 42.3 & 57.7 \\
& CPMF                & 61.1 & 54.5 & {78.3} & 45.8 & 65.7 & 45.8 & 48.6 & 58.2 & 58.2 & 51.4 & 65.1 & 50.4 & 52.9 & 54.5 & 57.3 \\
& Reg3D-AD            & 59.9 & 59.9 & {68.8} & 58.9 & {74.1} & 54.8 & 60.2 & 40.5 & 51.1 & {75.5} & 62.4 & {88.1} & {81.1} & {69.4} & {66.8} \\
& IMRNet              & {76.5} & 74.2 & 60.5 & 68.1 & 69.9 & 53.5 & {68.5} & 61.4 & 40.1 & 52.4 & {68.2} & 59.3 & 63.5 & 69.1 & 65.0 \\
& PO3AD               & {87.1} & {81.0} & 66.3 & {78.3} & 69.2 & {95.5} & \underline{88.2} & \underline{97.8} & \underline{88.4} & {90.2} & \underline{93.7} & \underline{98.2} & \underline{95.0} & \underline{95.2} & \underline{89.8} \\
& PASDF               & {95.9} & {95.1} & \underline{86.5} & {88.4} & \textbf{90.2} & {94.4} & {79.7} & {95.6} & {86.8} & {89.9} & {91.5} & {95.9} & {90.9} & {86.3} & {89.7} \\
& Reg2Inv             & \textbf{98.2} & \textbf{99.2} & 63.2 & \textbf{91.8} & 86.9 & \textbf{98.0} & {70.5} & \textbf{99.7} & {84.4} & \underline{92.7} & {87.9} & {86.3} & {93.4} & \textbf{97.1} & {88.2} \\
\midrule[1.2pt]
\multirow{2}{*}{\textbf{Unified}}
& MC3D-AD             & {84.8} & {85.4} & {65.8} & {52.4} & {53.7} & {87.0} & {72.6} & {82.2} & {79.7} & {78.6} & {64.7} & {61.6} & {88.5} & {77.6} & {75.9} \\
& \cellcolor{pubcolor!25}\textbf{GRIM (Ours)} & \cellcolor{pubcolor!25}\underline{97.5} & \cellcolor{pubcolor!25}\underline{97.3} & \cellcolor{pubcolor!25}\textbf{93.5} & \cellcolor{pubcolor!25}\underline{90.1} & \cellcolor{pubcolor!25}\underline{88.1} & \cellcolor{pubcolor!25}\underline{97.9} & \cellcolor{pubcolor!25}\textbf{96.0} & \cellcolor{pubcolor!25}{95.8} & \cellcolor{pubcolor!25}\textbf{97.4} & \cellcolor{pubcolor!25}\textbf{97.2} & \cellcolor{pubcolor!25}\textbf{94.0} & \cellcolor{pubcolor!25}\textbf{98.9} & \cellcolor{pubcolor!25}\textbf{97.4} & \cellcolor{pubcolor!25}{94.1} & \cellcolor{pubcolor!25}\textbf{94.5} \\
\bottomrule
\end{tabular}
}
\label{tab:table9}
\end{table*}

\begin{table*}[!ht]
\caption{Comparison of object-level AUROC (\%) across all categories on Real3D-AD.}
\resizebox{\textwidth}{!}{
\begin{tabular}{c|l|cccccccccccc|c}
\toprule
\multicolumn{15}{c}{O-AUROC($\uparrow$)} \\ \midrule
\multicolumn{2}{c|}{Method} & Airplane & Car & Candybar & Chicken & Diamond & Duck & Fish & Gemstone & Seahorse & Shell & Starfish & Toffees & Mean \\ 
\midrule
\multirow{14}{*}{\shortstack{Category-\\Specific}}
& BTF(Raw)            & 73.0 & 64.7 & 53.9 & 78.9 & 70.7 & 69.1 & 60.2 & {68.6} & 59.6 & 39.6 & 53.0 & 70.3 & 63.5 \\
& BTF(FPFH)           & 52.0 & 56.0 & 63.0 & 43.2 & 54.5 & 78.4 & 54.9 & 64.8 & {77.9} & 75.4 & 57.5 & 46.2 & 60.3 \\
& M3DM                & 43.4 & 54.1 & 55.2 & 68.3 & 60.2 & 43.3 & 54.0 & 64.4 & 49.5 & 69.4 & 55.1 & 45.0 & 59.4 \\
& PatchCore(FPFH)     & \textbf{88.2} & 59.0 & 54.1 & {83.7} & 57.4 & 54.6 & 67.5 & 37.0 & 50.5 & 58.9 & 44.1 & 56.5 & 59.3 \\
& PatchCore(PointMAE) & 72.6 & 49.8 & 66.3 & 82.7 & 78.3 & 48.9 & 63.0 & 37.4 & 53.9 & 50.1 & 51.9 & 58.5 & 59.4 \\
& CPMF                & 70.1 & 55.1 & 55.2 & 50.4 & 52.3 & 58.2 & 55.8 & 58.9 & 72.9 & 65.3 & 70.0 & 39.0 & 58.6 \\
& IMRNet              & 76.2 & 71.1 & 75.5 & 78.0 & 90.5 & 51.7 & 88.0 & 67.4 & 60.4 & 66.5 & 67.4 & 77.4 & 72.5 \\
& Reg3D-AD            & 71.6 & 69.7 & 68.5 & {85.2} & 90.0 & 58.4 & 91.5 & 41.7 & 76.2 & 58.3 & 50.6 & 82.7 & 70.4 \\
& Group3AD            & 74.4 & {72.8} & {84.7} & 78.6 & {93.2} & 67.9 & \underline{97.6} & 53.9 & \underline{84.1} & 58.5 & 56.2 & {79.6} & 75.1 \\
& R3D-AD              & 77.2 & 69.6 & 71.3 & 71.4 & 68.5 & \textbf{90.9} & 69.2 & 66.5 & 72.0 & {84.0} & {70.1} & 70.3 & 73.4 \\
& ISMP                & \underline{85.8} & {73.1} & {85.2} & 71.4 & {94.8} & 71.2 & {94.5} & 46.8 & 72.9 & 62.3 & 66.0 & \underline{84.2} & {76.7} \\
& PO3AD               & 80.4 & 65.4 & 78.5 & 68.6 & 80.1 & {82.0} & 85.9 & {69.3} & 75.6 & {80.0} & {75.8} & 77.1 & {76.5} \\
& PASDF               & 62.8 & \textbf{95.9} & 78.8 & 73.9 & 89.4 & {65.8} & \textbf{98.9} & {63.4} & \textbf{100.0} & \underline{85.0} & {61.7} & \textbf{86.6} & \underline{80.2} \\
& Reg2Inv             & 81.8 & 75.8 & \textbf{100.0} & \textbf{94.4} & \textbf{100.0} & {75.0} & 67.2 & \underline{73.5} & 53.2 & {69.2} & \underline{84.1} & 62.6 & {78.0} \\
\midrule[1.2pt]
\multirow{2}{*}{\textbf{Unified}}
& MC3D-AD             & {85.0} & {74.9} & {83.0} & {71.5} & {95.5} & \underline{83.1} & {86.5} & {56.0} & {71.6} & {80.3} & {76.6} & {73.8} & {78.2} \\
& \cellcolor{pubcolor!25}\textbf{GRIM (Ours)} & \cellcolor{pubcolor!25}{80.5} & \cellcolor{pubcolor!25}\underline{79.9} & \cellcolor{pubcolor!25}\underline{93.8} & \cellcolor{pubcolor!25}\underline{87.2} & \cellcolor{pubcolor!25}\underline{99.1} & \cellcolor{pubcolor!25}{81.5} & \cellcolor{pubcolor!25}{91.1} & \cellcolor{pubcolor!25}\textbf{80.0} & \cellcolor{pubcolor!25}{82.6} & \cellcolor{pubcolor!25}\textbf{96.5} & \cellcolor{pubcolor!25}\textbf{91.2} & \cellcolor{pubcolor!25}{82.6} & \cellcolor{pubcolor!25}\textbf{87.2} \\
\bottomrule
\end{tabular}
}
\label{tab:table10}
\end{table*}

\renewcommand{\arraystretch}{1.08}

\begin{table*}[!ht]
\caption{Comparison of point-level AUROC (\%) across all categories on Real3D-AD.}
\resizebox{\textwidth}{!}{
\begin{tabular}{c|l|cccccccccccc|c}
\toprule
\multicolumn{15}{c}{P-AUROC($\uparrow$)} \\ \midrule
\multicolumn{2}{c|}{Method} & Airplane & Car & Candybar & Chicken & Diamond & Duck & Fish & Gemstone & Seahorse & Shell & Starfish & Toffees & Mean \\ 
\midrule
\multirow{12}{*}{\shortstack{Category-\\Specific}}
& BTF(Raw)            & 56.4 & 64.7 & 73.5 & 60.9 & 56.3 & 60.1 & 51.4 & 59.7 & 52.0 & 48.9 & 39.2 & 62.3 & 57.1 \\
& BTF(FPFH)           & {73.8} & 70.8 & {86.4} & 73.5 & {88.2} & {87.5} & 70.9 & {89.1} & 51.2 & 57.1 & 50.1 & 81.5 & 73.3 \\
& M3DM                & 54.7 & 60.2 & 67.9 & 67.8 & 60.8 & 66.7 & 60.6 & 67.4 & 56.0 & 73.8 & 53.2 & 68.2 & 62.0 \\
& PatchCore(FPFH)     & 56.2 & {75.4} & 78.0 & 42.9 & 82.8 & 26.4 & 82.9 & \underline{91.0} & 73.9 & 73.9 & 60.6 & 74.7 & 68.2 \\
& PatchCore(PointMAE) & 56.9 & 60.9 & 62.7 & 72.9 & 71.8 & 52.8 & 71.7 & 44.4 & 63.3 & 70.9 & 58.0 & 58.0 & 62.0 \\
& CPMF                & 61.8 & {83.6} & 73.4 & 55.9 & 75.3 & 71.9 & \textbf{98.8} & 44.9 & \textbf{96.2} & 72.5 & {80.0} & \textbf{95.9} & {75.8} \\
& Reg3D-AD            & 63.1 & 71.8 & 72.4 & 67.6 & 83.5 & 50.3 & 82.6 & 54.5 & 81.7 & {81.1} & 61.7 & 75.9 & 70.5 \\
& Group3AD            & 63.6 & 74.5 & 73.8 & {75.9} & 86.2 & 63.1 & 83.6 & 56.4 & {82.7} & 79.8 & 62.5 & 80.3 & 73.5 \\
& ISMP                & {75.3} & {83.6} & {90.7} & {79.8} & {92.6} & {87.6} & {88.6} & 85.7 & 81.3 & {83.9} & 64.1 & {89.5} & {83.6} \\
& PO3AD               & 71.5 & 57.4 & 76.3 & 56.3 & 61.4 & 64.4 & 87.7 & 53.0 & 58.5 & 54.2 & {65.9} & 72.9 & 65.0 \\
& PASDF               & \underline{77.7} & 80.2 & 54.6 & 76.8 & 69.9 & 78.2 & 83.7 & 65.4 & \underline{88.7} & 64.8 & {70.3} & 83.8 & 74.5 \\
& Reg2Inv             & \textbf{92.3} & \textbf{94.4} & \textbf{96.9} & \textbf{91.0} & \textbf{97.9} & \underline{93.7} & 84.6 & 90.7 & 64.5 & \underline{90.6} & \underline{84.0} & 73.7 & \underline{87.8} \\
\midrule[1.2pt]
\multirow{2}{*}{\textbf{Unified}}
& MC3D-AD             & {62.8} & {81.9} & {91.0} & {64.0} & {94.2} & {82.2} & {93.2} & {45.8} & {65.9} & {77.8} & {69.0} & {93.4} & {76.8} \\
& \cellcolor{pubcolor!25}\textbf{GRIM (Ours)} & \cellcolor{pubcolor!25}{75.9} & \cellcolor{pubcolor!25}\underline{87.6} & \cellcolor{pubcolor!25}\underline{94.4} & \cellcolor{pubcolor!25}\underline{90.7} & \cellcolor{pubcolor!25}\underline{97.6} & \cellcolor{pubcolor!25}\textbf{94.1} & \cellcolor{pubcolor!25}\underline{97.3} & \cellcolor{pubcolor!25}\textbf{92.5} & \cellcolor{pubcolor!25}{87.0} & \cellcolor{pubcolor!25}\textbf{96.2} & \cellcolor{pubcolor!25}\textbf{87.4} & \cellcolor{pubcolor!25}\underline{94.5} & \cellcolor{pubcolor!25}\textbf{91.3} \\
\bottomrule
\end{tabular}
}
\label{tab:table11}
\end{table*}

\begin{table*}[!ht]
\caption{Comparison of object-level AUROC (\%) across all categories on Real3D-AD under the cross-domain setting. Models are trained on Anomaly-ShapeNet and tested on Real3D-AD.}
\resizebox{\textwidth}{!}{
\begin{tabular}{l|cccccccccccc|c}
\toprule
\multicolumn{14}{c}{O-AUROC($\uparrow$)} \\ \midrule
Method & Airplane & Car & Candybar & Chicken & Diamond & Duck & Fish & Gemstone & Seahorse & Shell & Starfish & Toffees & Mean \\ 
\midrule
MC3D-AD              & 42.8 & 48.2 & 58.8 & 50.5 & 67.0 & 63.3 & 49.5 & 57.1 & 57.8 & 59.4 & 65.6 & 45.4 & 55.4 \\
\rowcolor{pubcolor!25}
\textbf{GRIM (Ours)}        & \textbf{78.0} & \textbf{70.2} & \textbf{76.4} & \textbf{81.2} & \textbf{96.2} & \textbf{85.9} & \textbf{81.6} & \textbf{81.0} & \textbf{92.8} & \textbf{97.1} & \textbf{88.0} & \textbf{74.6} & \textbf{83.6} \\
\bottomrule
\end{tabular}
}
\label{tab:table12}
\end{table*}

\begin{table*}[!ht]
\caption{Comparison of point-level AUROC (\%) across all categories on Real3D-AD under the cross-domain setting. Models are trained on Anomaly-ShapeNet and tested on Real3D-AD.}
\resizebox{\textwidth}{!}{
\begin{tabular}{l|cccccccccccc|c}
\toprule
\multicolumn{14}{c}{P-AUROC($\uparrow$)} \\ \midrule
Method & Airplane & Car & Candybar & Chicken & Diamond & Duck & Fish & Gemstone & Seahorse & Shell & Starfish & Toffees & Mean \\ 
\midrule
MC3D-AD              & 38.5 & 25.0 & 32.1 & 44.1 & 37.4 & 36.5 & 35.9 & 41.2 & 36.3 & 53.8 & 50.1 & 29.1 & 38.3 \\
\rowcolor{pubcolor!25}
\textbf{GRIM (Ours)}        & \textbf{76.3} & \textbf{87.0} & \textbf{92.0} & \textbf{84.6} & \textbf{93.9} & \textbf{90.1} & \textbf{95.2} & \textbf{92.2} & \textbf{86.1} & \textbf{97.3} & \textbf{89.6} & \textbf{91.3} & \textbf{89.6} \\
\bottomrule
\end{tabular}
}
\label{tab:table13}
\end{table*}

\begin{table*}[!ht]
\caption{Comparison of object-level AUROC (\%) across all categories on Anomaly-ShapeNet under the cross-domain setting. Models are trained on Real3D-AD and tested on Anomaly-ShapeNet.}
\resizebox{\textwidth}{!}{
\begin{tabular}{l|cccccccccccccccccccc}
\toprule
\multicolumn{21}{c}{O-AUROC($\uparrow$)}\\
\midrule
Method & ashtray0 & bag0 & bottle0 & bottle1 & bottle3 & bowl0 & bowl1 & bowl2 & bowl3 & bowl4 & bowl5 & bucket0 & bucket1 & cap0 & cap3 & cap4 & cap5 & cup0 & cup1 & eraser0 \\
\midrule
MC3D-AD             & 62.9 & 68.1 & 77.6 & 80.0 & 78.1 & \textbf{97.8} & 73.7 & 75.2 & 74.1 & 78.5 & 76.1 & 74.6 & 75.2 & 61.9 & 89.8 & 87.0 & 80.4 & 74.8 & 86.2 & \textbf{80.0} \\
\rowcolor{pubcolor!25}
\textbf{GRIM (Ours)}       & \textbf{96.2} & \textbf{79.0} & \textbf{85.7} & \textbf{82.8} & \textbf{99.7} & 97.0 & \textbf{91.1} & \textbf{100.0} & \textbf{93.7} & \textbf{91.5} & \textbf{95.4} & \textbf{94.0} & \textbf{85.1} & \textbf{87.0} & \textbf{100.0} & \textbf{100.0} & \textbf{95.4} & \textbf{91.0} & \textbf{91.4} & \textbf{80.0} \\
\bottomrule
\end{tabular}
}
\resizebox{\textwidth}{!}{
\begin{tabular}{l|cccccccccccccccccccc|c}
\toprule
Method & headset0 & headset1 & helmet0 & helmet1 & helmet2 & helmet3 & jar & phone & shelf0 & tap0 & tap1 & vase0 & vase1 & vase2 & vase3 & vase4 & vase5 & vase7 & vase8 & vase9 & Mean \\
\midrule
MC3D-AD             & 86.2 & 81.0 & 77.7 & 79.5 & 76.8 & \textbf{92.1} & \textbf{93.3} & 85.2 & 65.5 & 67.9 & 71.5 & 86.3 & 70.0 & 90.5 & 80.3 & 85.2 & 81.0 & 78.1 & 70.3 & 63.3 & 78.3 \\
\rowcolor{pubcolor!25}
\textbf{GRIM (Ours)}       & \textbf{95.6} & \textbf{95.7} & \textbf{84.9} & \textbf{96.2} & \textbf{94.5} & 84.8 & 88.6 & \textbf{92.4} & \textbf{82.0} & \textbf{98.5} & \textbf{97.0} & \textbf{97.9} & \textbf{78.6} & \textbf{94.8} & \textbf{85.2} & \textbf{99.4} & \textbf{92.4} & \textbf{80.0} & \textbf{92.7} & \textbf{98.8} & \textbf{91.7} \\
\bottomrule
\end{tabular}
}
\label{tab:table14}
\vspace{-4mm}
\end{table*}

\begin{table*}[t!]
\caption{Comparison of point-level AUROC (\%) across all categories on Anomaly-ShapeNet under the cross-domain setting. Models are trained on Real3D-AD and tested on Anomaly-ShapeNet.}
\resizebox{\textwidth}{!}{
\begin{tabular}{l|ccccccccccccc}
\toprule
\multicolumn{14}{c}{P-AUROC($\uparrow$)}\\
\midrule
Method & ashtray0 & bag0 & bottle0 & bottle1 & bottle3 & bowl0 & bowl1 & bowl2 & bowl3 & bowl4 & bowl5 & bucket0 & bucket1 \\
\midrule
MC3D-AD      & 48.6 & 51.3 & 56.2 & 62.7 & 48.5 & 49.5 & 43.4 & 48.6 & 43.1 & 50.3 & 43.6 & 55.3 & 47.3 \\
\rowcolor{pubcolor!25}
\textbf{GRIM (Ours)}     & \textbf{82.1} & \textbf{85.6} & \textbf{86.5} & \textbf{78.6} & \textbf{84.9} & \textbf{75.6} & \textbf{61.7} & \textbf{84.6} & \textbf{77.0} & \textbf{79.3} & \textbf{81.6} & \textbf{77.1} & \textbf{80.8} \\
\bottomrule
\end{tabular}
}

\resizebox{\textwidth}{!}{
\begin{tabular}{l|cccccccccccccc}
\toprule
Method & cap0 & cap3 & cap4 & cap5 & cup0 & cup1 & eraser0 & headset0 & headset1 & helmet0 & helmet1 & helmet2 & helmet3 \\
\midrule
MC3D-AD      & 46.8 & 54.2 & 48.7 & 48.3 & 54.0 & 36.5 & 56.5 & 47.1 & 55.0 & 49.6 & 37.0 & 53.7 & 64.8 \\
\rowcolor{pubcolor!25}
\textbf{GRIM (Ours)}     & \textbf{87.6} & \textbf{86.7} & \textbf{81.6} & \textbf{86.6} & \textbf{87.5} & \textbf{81.0} & \textbf{84.9} & \textbf{78.8} & \textbf{79.6} & \textbf{86.0} & \textbf{75.3} & \textbf{74.8} & \textbf{74.6} \\
\bottomrule
\end{tabular}
}

\resizebox{\textwidth}{!}{
\begin{tabular}{l|cccccccccccccc|c}
\toprule
Method & jar & phone & shelf0 & tap0 & tap1 & vase0 & vase1 & vase2 & vase3 & vase4 & vase5 & vase7 & vase8 & vase9 & Mean \\
\midrule
MC3D-AD       & 36.3 & 52.2 & 51.8 & 44.1 & 46.5 & 46.8 & \textbf{65.9} & 44.0 & 46.8 & 48.1 & 44.2 & 41.4 & 47.0 & 35.3 & 48.8 \\
\rowcolor{pubcolor!25}
\textbf{GRIM (Ours)}     & \textbf{77.5} & \textbf{85.8} & \textbf{82.0} & \textbf{76.2} & \textbf{75.6} & \textbf{80.9} & 65.7 & \textbf{64.3} & \textbf{81.2} & \textbf{78.7} & \textbf{72.1} & \textbf{80.6} & \textbf{59.0} & \textbf{74.6} & \textbf{78.9} \\
\bottomrule
\end{tabular}
}
\label{tab:table15}
\end{table*}

\begin{table*}[t!]
\caption{Comparison of object-level AUPR (\%) across all categories on Anomaly-ShapeNet.}
\resizebox{\textwidth}{!}{
\begin{tabular}{c|l|ccccccccccccc}
\toprule
\multicolumn{15}{c}{O-AUPR($\uparrow$)}\\
\midrule
\multicolumn{2}{c|}{Method} & ashtray0 & bag0 & bottle0 & bottle1 & bottle3 & bowl0 & bowl1 & bowl2 & bowl3 & bowl4 & bowl5 & bucket0 & bucket1 \\
\midrule
\multirow{9}{*}{\shortstack{Category-\\Specific}}
& BTF(Raw)            & 57.8 & 45.8 & 46.6 & 57.3 & 54.3 & 58.8 & 46.4 & 57.6 & {65.4} & 60.1 & 61.5 & 65.2 & 62.0 \\
& BTF(FPFH)           & 65.1 & 55.1 & 64.4 & 62.5 & 60.2 & 57.6 & {64.8} & 51.5 & 49.9 & 63.2 & {69.9} & 48.3 & 64.8 \\
& M3DM                & 63.2 & 64.2 & {76.3} & 67.4 & 45.1 & 52.5 & 51.5 & 63.0 & 63.5 & 57.1 & 60.1 & 60.9 & 50.7 \\
& PatchCore(FPFH)     & 44.5 & 60.8 & 61.5 & 67.7 & 57.9 & 54.8 & 54.5 & 61.1 & 62.0 & 57.5 & 54.1 & 60.4 & 56.5 \\
& PatchCore(PointMAE) & {67.9} & 60.1 & 54.5 & 64.5 & {65.1} & 56.2 & 61.1 & 45.6 & 55.6 & 60.1 & 58.5 & 54.1 & 64.2 \\
& CPMF                & 45.3 & 65.5 & 58.8 & 59.2 & 50.5 & {77.5} & 62.1 & 60.1 & 41.8 & {68.3} & 68.5 & {66.2} & 50.1 \\
& Reg3D-AD            & 58.8 & 60.8 & 63.2 & 69.5 & 47.4 & 49.4 & 51.5 & 49.5 & 44.1 & 62.4 & 55.5 & 63.2 & 71.4 \\
& IMRNet              & 61.2 & {66.5} & 55.8 & {70.2} & 64.8 & 48.1 & 50.4 & {68.1} & 61.4 & 63.0 & 65.2 & 57.8 & {73.2} \\
& PO3AD               & \underline{99.9} & \underline{80.9} & \underline{92.7} & \underline{95.9} & \textbf{96.2} & \underline{94.6} & \underline{90.5} & \underline{88.8} & \underline{92.7} & \textbf{98.5} & \underline{90.4} & \underline{92.3} & \underline{88.2} \\
\midrule[1.2pt]
\multirow{1}{*}{\textbf{Unified}}
& \cellcolor{pubcolor!25}\textbf{GRIM (Ours)} & \cellcolor{pubcolor!25}\textbf{100.0} & \cellcolor{pubcolor!25}\textbf{91.6} & \cellcolor{pubcolor!25}\textbf{95.6} & \cellcolor{pubcolor!25}\textbf{100.0} & \cellcolor{pubcolor!25}\underline{92.7} & \cellcolor{pubcolor!25}\textbf{95.1} & \cellcolor{pubcolor!25}\textbf{100.0} & \cellcolor{pubcolor!25}\textbf{98.4} & \cellcolor{pubcolor!25}\textbf{95.3} & \cellcolor{pubcolor!25}\textbf{98.5} & \cellcolor{pubcolor!25}\textbf{100.0} & \cellcolor{pubcolor!25}\textbf{98.0} & \cellcolor{pubcolor!25}\textbf{100.0} \\
\bottomrule
\end{tabular}
}
\resizebox{\textwidth}{!}{
\begin{tabular}{c|l|ccccccccccccc}
\toprule
\multicolumn{2}{c|}{Method} & cap0 & cap3 & cap4 & cap5 & cup0 & cup1 & eraser0 & headset0 & headset1 & helmet0 & helmet1 & helmet2 & helmet3 \\
\midrule
\multirow{9}{*}{\shortstack{Category-\\Specific}}
& BTF(Raw)            & 65.9 & 61.2 & 51.5 & 65.3 & 60.1 & 70.1 & 42.5 & 37.9 & 51.5 & 55.9 & 38.8 & 61.5 & 52.6 \\
& BTF(FPFH)           & 61.8 & 57.9 & 54.5 & 59.3 & 58.5 & 65.1 & 71.9 & 53.1 & 52.3 & 56.8 & {72.1} & 58.8 & 56.4 \\
& M3DM                & 56.4 & 65.2 & 47.7 & 64.2 & 57.0 & {75.2} & 62.5 & 63.2 & 62.3 & 52.8 & 62.7 & {63.6} & 45.8 \\
& PatchCore(FPFH)     & 58.5 & 45.7 & 65.5 & 72.5 & 60.4 & 58.6 & 58.4 & {70.1} & 60.1 & 52.5 & 63.0 & 47.5 & 49.4 \\
& PatchCore(PointMAE) & 56.1 & 58.3 & {72.1} & 54.2 & 64.2 & 71.0 & {80.1} & 51.5 & 42.3 & 63.3 & 57.1 & 49.6 & 61.1 \\
& CPMF                & 60.1 & 54.1 & 64.5 & 69.7 & {64.7} & 60.9 & 54.4 & 60.2 & 61.9 & 33.3 & 50.1 & 47.7 & {64.5} \\
& Reg3D-AD            & 69.3 & {71.1} & 62.3 & {77.0} & 53.1 & 63.8 & 42.4 & 53.8 & 61.7 & 60.0 & 38.1 & 61.8 & 46.8 \\
& IMRNet              & {71.1} & 70.2 & 65.8 & 50.2 & 45.5 & 62.7 & 59.9 & {70.1} & {65.6} & {69.7} & 61.5 & 60.2 & 57.5 \\
& PO3AD               & \underline{84.1} & \underline{90.6} & \underline{87.6} & \underline{80.1} & \underline{87.9} & \underline{87.0} & \textbf{99.5} & \underline{76.5} & \underline{91.4} & \underline{86.4} & \underline{96.1} & \underline{93.4} & \underline{84.9} \\
\midrule[1.2pt]
\multirow{1}{*}{\textbf{Unified}}
& \cellcolor{pubcolor!25}\textbf{GRIM (Ours)} & \cellcolor{pubcolor!25}\textbf{100.0} & \cellcolor{pubcolor!25}\textbf{99.7} & \cellcolor{pubcolor!25}\textbf{96.3} & \cellcolor{pubcolor!25}\textbf{96.0} & \cellcolor{pubcolor!25}\textbf{100.0} & \cellcolor{pubcolor!25}\textbf{100.0} & \cellcolor{pubcolor!25}\underline{97.9} & \cellcolor{pubcolor!25}\textbf{98.4} & \cellcolor{pubcolor!25}\textbf{99.5} & \cellcolor{pubcolor!25}\textbf{97.8} & \cellcolor{pubcolor!25}{98.0} & \cellcolor{pubcolor!25}\textbf{99.8} & \cellcolor{pubcolor!25}\textbf{96.3} \\
\bottomrule
\end{tabular}
}
\resizebox{\textwidth}{!}{
\begin{tabular}{c|l|cccccccccccccc|c}
\toprule
\multicolumn{2}{c|}{Method} & jar & phone & shelf0 & tap0 & tap1 & vase0 & vase1 & vase2 & vase3 & vase4 & vase5 & vase7 & vase8 & vase9 & Mean \\
\midrule
\multirow{9}{*}{\shortstack{Category-\\Specific}}
& BTF(Raw)            & 42.8 & 61.3 & 62.4 & 53.5 & 59.4 & 56.2 & 44.1 & 41.3 & {71.7} & 42.8 & 61.5 & 54.7 & 41.6 & 48.2 & 54.9 \\
& BTF(FPFH)           & 47.9 & {66.2} & 61.1 & 61.0 & 57.5 & 64.1 & 65.5 & 56.9 & 65.2 & 58.7 & 47.2 & 59.2 & 62.4 & 63.8 & 59.8 \\
& M3DM                & 55.5 & 46.4 & 66.5 & {72.2} & 63.8 & {78.8} & 65.2 & 61.5 & 55.1 & 52.6 & 63.3 & 64.8 & 46.3 & 65.1 & 60.3 \\
& PatchCore(FPFH)     & 49.9 & 33.2 & 50.4 & 71.2 & 68.4 & 64.5 & 62.3 & {80.1} & 48.1 & {77.7} & 51.5 & 62.1 & 51.5 & {66.0} & 58.8 \\
& PatchCore(PointMAE) & 46.3 & 65.2 & 54.3 & 71.2 & 54.2 & 54.8 & 57.2 & 71.1 & 45.5 & 58.6 & 58.5 & {65.2} & 65.5 & 63.4 & 59.5 \\
& CPMF                & 61.8 & 65.5 & \underline{68.1} & 63.9 & 69.7 & 63.2 & 64.5 & 63.2 & 58.8 & 65.5 & 51.8 & 43.2 & {67.3} & 61.8 & 59.7 \\
& Reg3D-AD            & 60.1 & 61.4 & 67.5 & 67.6 & 59.9 & 61.5 & 46.8 & 64.1 & 65.1 & 50.5 & 58.8 & 45.5 & 62.9 & 57.4 & 58.4 \\
& IMRNet              & {76.0} & 55.2 & 62.5 & 40.1 & \underline{79.6} & 57.3 & {72.5} & 65.5 & 70.8 & 52.8 & {65.4} & 60.1 & 63.9 & 46.2 & {62.1} \\
& PO3AD               & \underline{91.5} & \underline{80.3} & {68.0} & \underline{85.6} & {70.9} & \underline{75.3} & \underline{78.9} & \textbf{96.3} & \underline{90.2} & \underline{82.4} & \underline{87.9} & \underline{97.1} & \underline{83.3} & \underline{90.4} & \underline{88.1} \\
\midrule[1.2pt]
\multirow{1}{*}{\textbf{Unified}}
& \cellcolor{pubcolor!25}\textbf{GRIM (Ours)} & \cellcolor{pubcolor!25}\textbf{100.0} & \cellcolor{pubcolor!25}\textbf{88.2} & \cellcolor{pubcolor!25}\textbf{93.0} & \cellcolor{pubcolor!25}\textbf{100.0} & \cellcolor{pubcolor!25}\textbf{81.7} & \cellcolor{pubcolor!25}\textbf{96.5} & \cellcolor{pubcolor!25}\textbf{91.8} & \cellcolor{pubcolor!25}\underline{95.9} & \cellcolor{pubcolor!25}\textbf{98.4} & \cellcolor{pubcolor!25}\textbf{100.0} & \cellcolor{pubcolor!25}\textbf{100.0} & \cellcolor{pubcolor!25}\textbf{99.5} & \cellcolor{pubcolor!25}\textbf{98.8} & \cellcolor{pubcolor!25}\textbf{93.6} & \cellcolor{pubcolor!25}\textbf{97.1} \\
\bottomrule
\end{tabular}
}
\label{tab:table16}
\end{table*}


\end{document}